\documentclass[a4paper,conference]{IEEEtran} %
\usepackage{multirow}
\usepackage{cite} 
  \usepackage[pdftex]{graphicx}
  \usepackage[caption=false,font=footnotesize]{subfig}

\usepackage[cmex10]{amsmath} 
\usepackage{amssymb}
\usepackage{booktabs} 

\usepackage{tabu}
\usepackage[linesnumbered,ruled,vlined]{algorithm2e}
\usepackage{etoolbox,lipsum}

\usepackage{stfloats}

\newcommand\T{\rule{0pt}{2.9ex}}       

\begin{document}



\title{On the Complexity of Object Detection on Real-world Public Transportation Images for Social Distancing Measurement}

\author{\IEEEauthorblockN{Nik Khadijah Nik Aznan\IEEEauthorrefmark{1}, John Brennan\IEEEauthorrefmark{2}, Daniel Bell\IEEEauthorrefmark{3}, Jennine Jonczyk\IEEEauthorrefmark{3} and Paul Watson\IEEEauthorrefmark{4}}

\IEEEauthorrefmark{1} Research Software Engineering, Newcastle University, Newcastle, UK
\\\IEEEauthorrefmark{2} National Innovation Centre for Data, Newcastle University, Newcastle, UK
\\\IEEEauthorrefmark{3} Urban Observatory, Newcastle University, Newcastle, UK
\\\IEEEauthorrefmark{4} School of Computing, Newcastle University, Newcastle, UK}



\maketitle

\begin{abstract}

  Social distancing in public spaces has become an essential aspect in helping to reduce the impact of the COVID-19 pandemic. Exploiting recent advances in machine learning, there have been many studies in the literature implementing social distancing via object detection through the use of surveillance cameras in public spaces. However, to date, there has been no study of social distance measurement on public transport. The public transport setting has some unique challenges, including some low-resolution images and camera locations that can lead to the partial occlusion of passengers, which make it challenging to perform accurate detection. Thus, in this paper, we investigate the challenges of performing accurate social distance measurement on public transportation. We benchmark several state-of-the-art object detection algorithms using real-world footage taken from the London Underground and bus network. The work highlights the complexity of performing social distancing measurement on images from current public transportation onboard cameras. Further, exploiting domain knowledge of expected passenger behaviour, we attempt to improve the quality of the detections using various strategies and show improvement over using vanilla object detection alone.
  
\end{abstract}

\IEEEpeerreviewmaketitle

\section{Introduction}
\label{sec:intro}

\begin{figure*}[!b]
  \centering
  \subfloat[Faster R-CNN]{\includegraphics[width=0.32\textwidth]{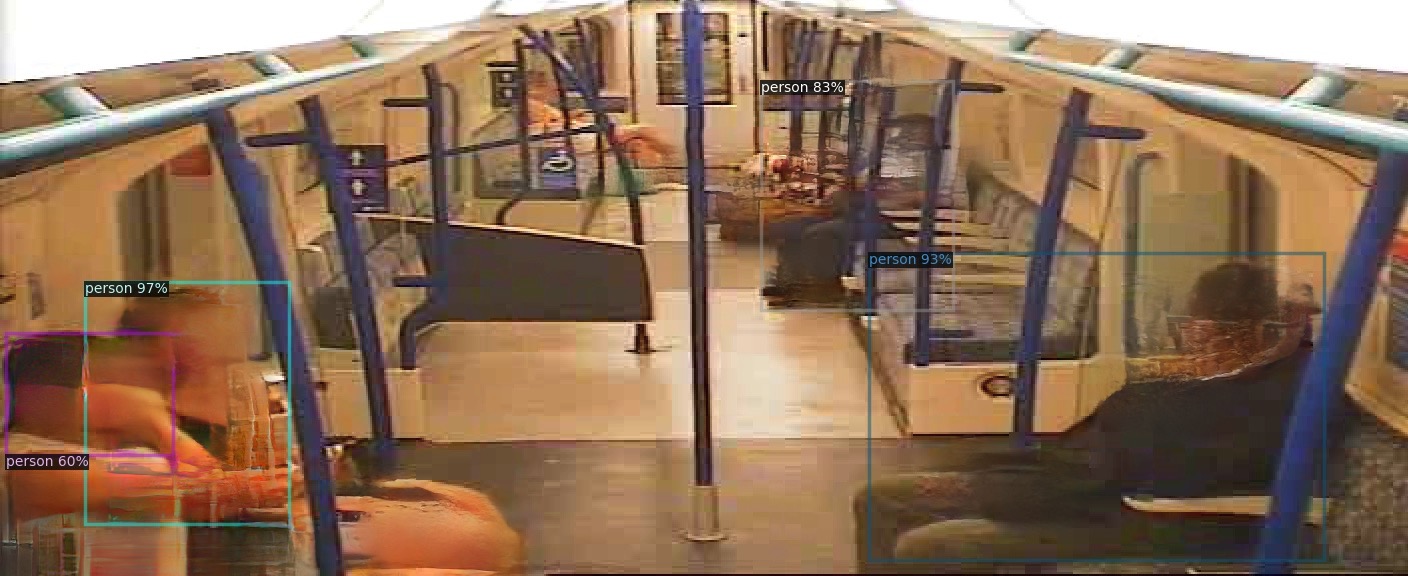}%
  \label{fig:compare_frcnn00}}
  \hfil
  \subfloat[Faster R-CNN]{\includegraphics[width=0.32\textwidth]{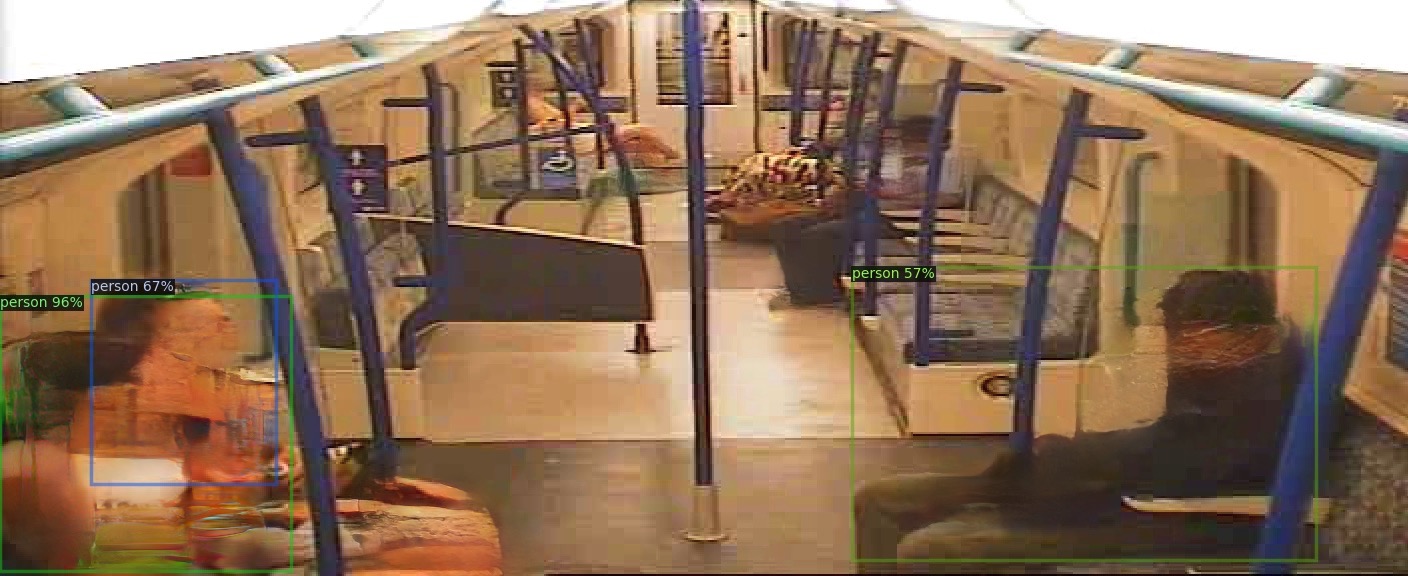}%
  \label{fig:compare_frcnn01}}
  \hfil
  \subfloat[Faster R-CNN]{\includegraphics[width=0.32\textwidth]{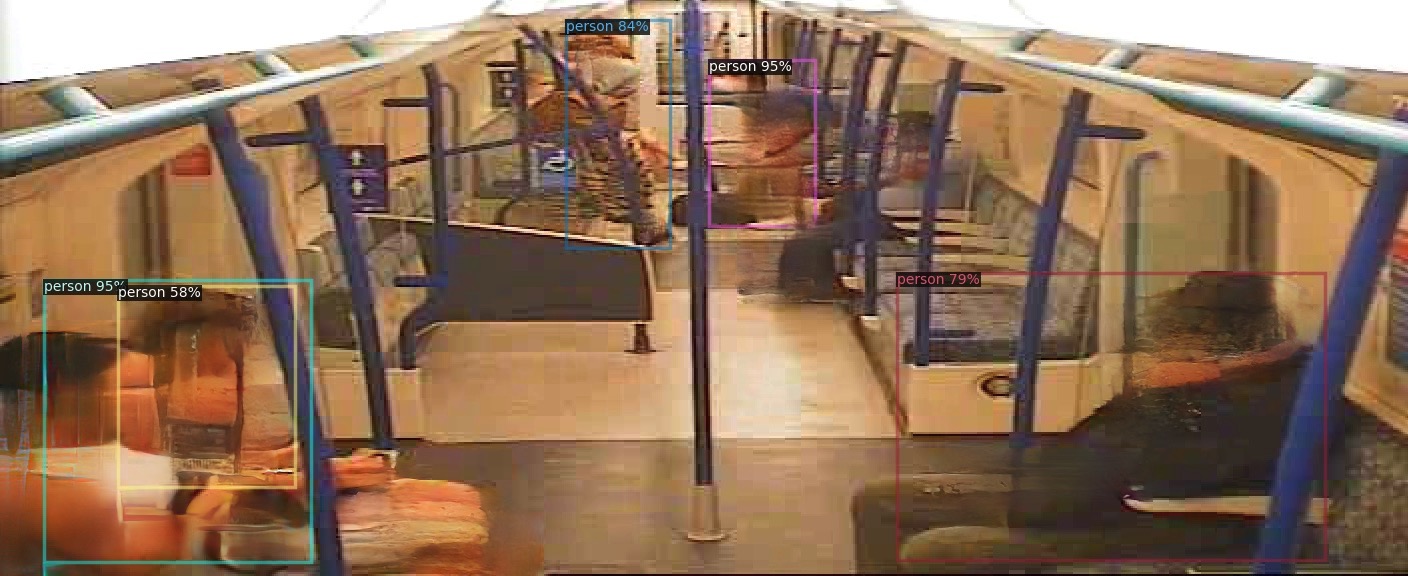}%
  \label{fig:compare_frcnn02}}

  \subfloat[Mask R-CNN]{\includegraphics[width=0.32\textwidth]{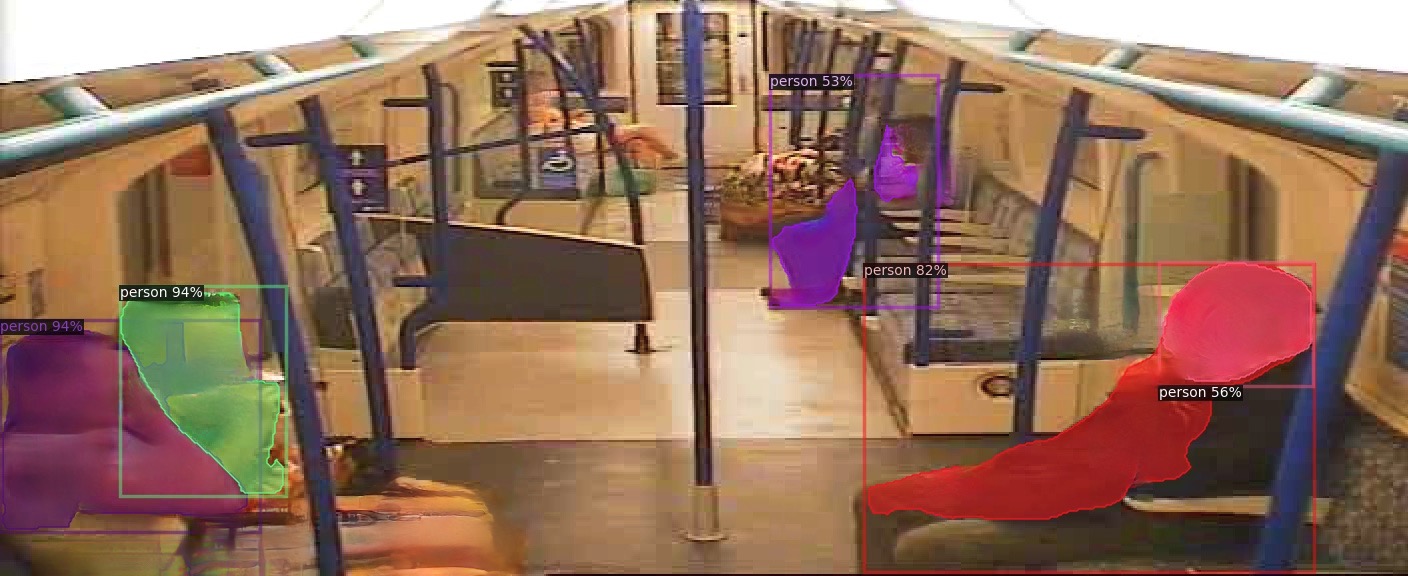}%
  \label{fig:compare_mrcnn00}}
  \hfil
  \subfloat[Mask R-CNN]{\includegraphics[width=0.32\textwidth]{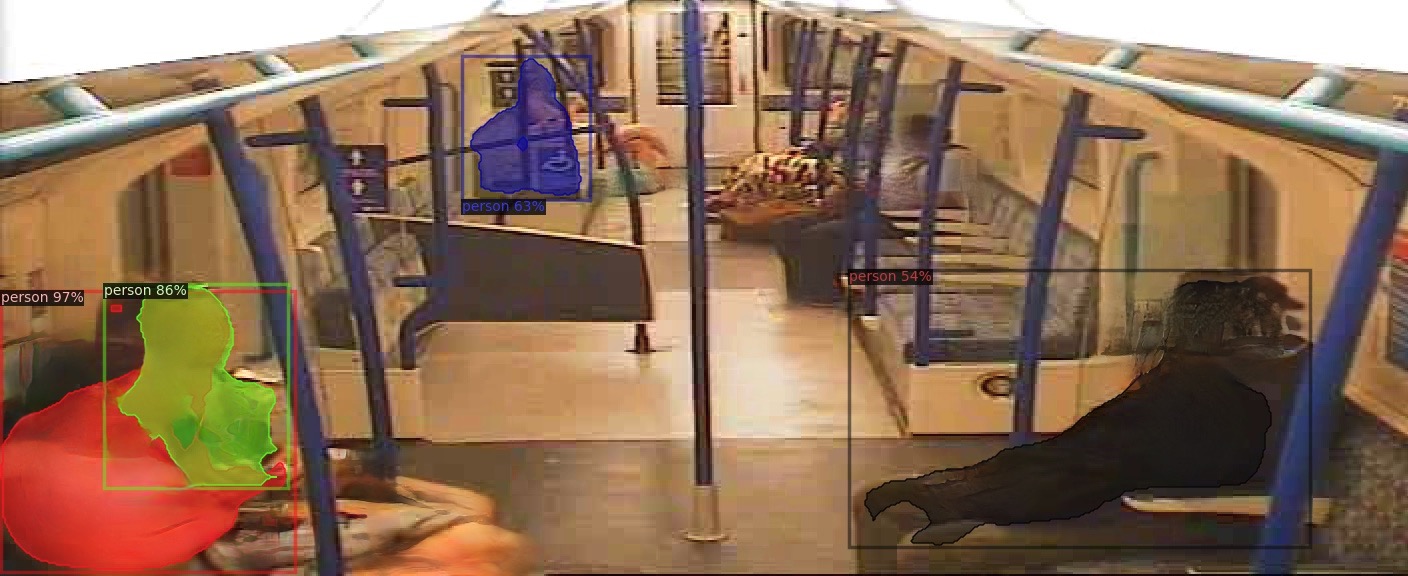}%
  \label{fig:compare_mrcnn01}}
  \hfil
  \subfloat[Mask R-CNN]{\includegraphics[width=0.32\textwidth]{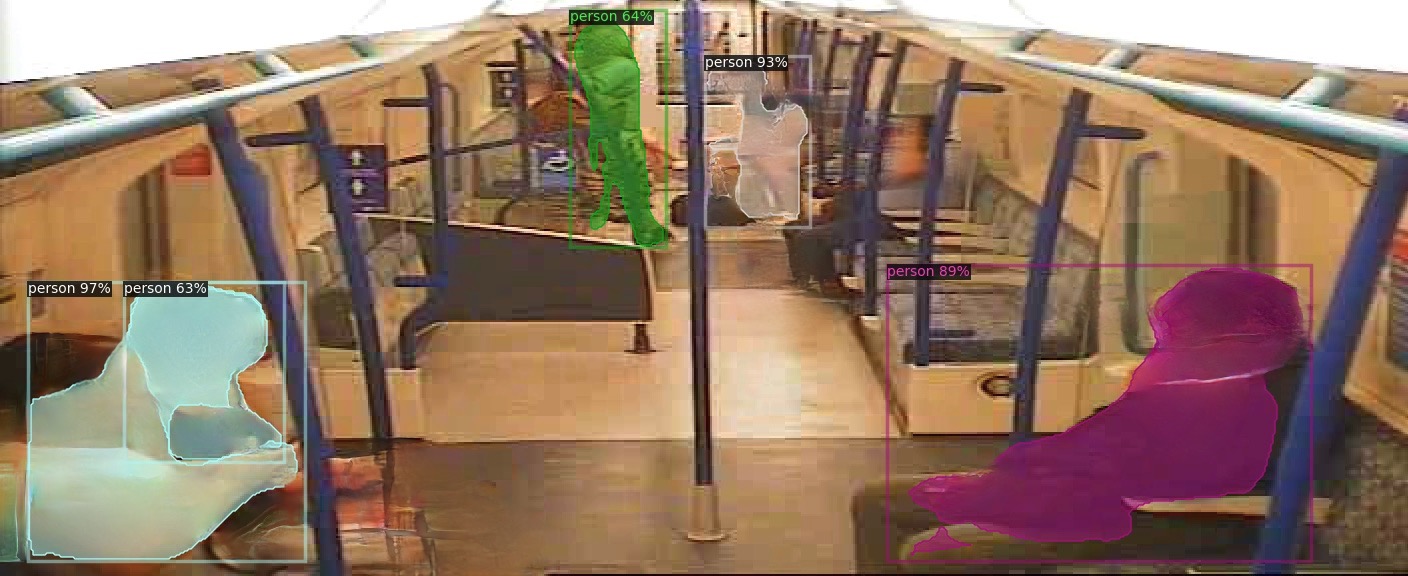}%
  \label{fig:compare_mrcnn02}}

  \subfloat[YOLOv3]{\includegraphics[width=0.32\textwidth]{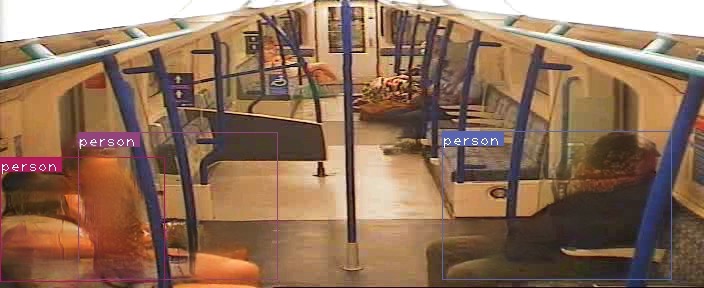}%
  \label{fig:compare_yolo00}}
  \hfil
  \subfloat[YOLOv3]{\includegraphics[width=0.32\textwidth]{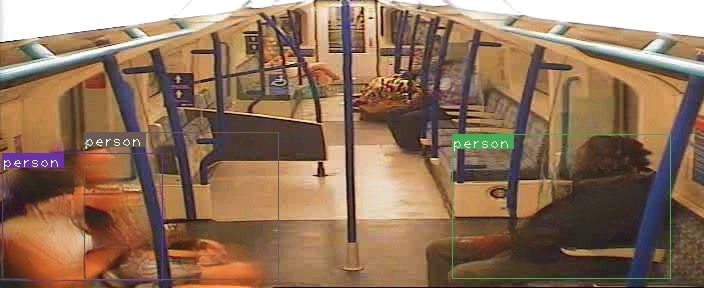}%
  \label{fig:compare_yolo01}}
  \hfil
  \subfloat[YOLOv3]{\includegraphics[width=0.32\textwidth]{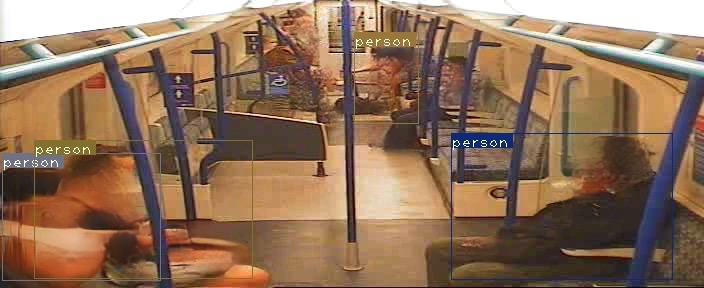}%
  \label{fig:compare_yolo02}}
  
  \caption{Preliminary passenger detection comparing three state-of-the-art object detection}
  \label{fig:Modelcompare}
\end{figure*}

There have been many studies on monitoring social distancing to provide safety measures in public spaces in recent years, primarily because of the COVID-19 pandemic. Social distancing refers to a distance maintained between one person to another to minimise physical contact~\cite{saponara2021implementing}.  One of the ways the virus spreads is through close contact between people; therefore, social distancing can reduce or slow down the spread~\cite{khataee2021effects}. By taking advantage of surveillance camera footage, social distancing monitoring systems have been developed that can be used to to either enforce social distancing in public spaces~\cite{varghese2021multimodal} or provide advice and guidelines on how effectively social distancing works~\cite{kumari2021deep, magoo2021deep, shalini2021social, punn2020monitoring, mercaldo2021proposal, hou2020social}. However, no study shows the efficacy of social distancing measurement on public transportation.

Public transportation is an essential facility for daily transportation. According to the United Kingdom Department of Transport, in the year 2019/2020, there were 4524 million passenger journeys on local bus services in Great Britain, 1745 million passenger journeys on the national rail network, and 1337 million passenger journeys on London underground~\cite{DfTRail}. These numbers showed that there were many journeys with a high-density of passengers and so automatic social distancing measurement would have been helpful. Whilst, social distancing on public transportation is critical as the virus spreads more easily in the enclosed spaces, other factors such as good ventilation on public transportation will also help with transmission mitigation and are highly recommended~\cite{morawska2020can}. However, social distancing is crucial when it is hard to achieve a good ventilation.

Convolutional Neural Networks (CNN) are a state-of-the-art machine learning approach that has been shown as a practical solution in object detection tasks within computer vision~\cite{zou2019object}. Object detection architectures such as region-based CNN (R-CNN) like Faster R-CNN , Mask R-CNN, and YOLO v3 (You Only Look Once version 3) provide solutions for detecting objects and their locations in an image~\cite{liu2020deep}. These large architectures are trained on massive benchmark datasets for multiple days to achieve their predictive performance. Despite advancements in computer vision, object detection tasks still face challenges when dealing with real-world images, especially where the images suffer from degradation and occlusion~\cite{liu2020deep}.

In order to accurately monitor social distancing on public transportation, ideally object detection should be able to detect the presence and position of every passenger on board by using the surveillance CCTV system on the vehicle. However, based on our observation of these images, the quality of the images is poor, with  very low-resolution and low frame-per-second when compared to the benchmark dataset used to train object detection algorithms such as MS COCO that has high quality and clean images~\cite{lin2014microsoft}. Additionally, the camera's physical position in the vehicle often means it is far from the location of the passenger seats and there are fittings such as hand rails and seats that prevent clear line of sight. This results in social distancing monitoring on public transportation being complex when compared to monitoring using CCTV in streets or malls. There are also possible false positives on the footage caused by either image degradation or the capture of partial images of people through the window and therefore outside the vehicle.  

In this study, we present three different object detection algorithms, Faster R-CNN, Mask R-CNN and YOLO v3, to investigate the performance of these algorithms in detecting passengers from the low-resolution onboard CCTV cameras from both busses and underground trains. We then explore using our domain knowledge on the expected behaviour of passengers to improve the accuracy. We also use density-based clustering of the detected passengers to examine the possibility of automatic social distancing measurement on public transportation. 

This paper highlights the complexity of achieving social distancing measurement from onboard low-resolution CCTV. To our knowledge, this study is the first to address such an issue on public transportation datasets. Using state-of-the-art computer vision algorithms such as Mask R-CNN, we believe performing social distance measurement in real-life will require some adjustment in terms of the hardware specification and the physical location of the surveillance camera. In summary, the main contributions of this study are:

\begin{itemize}
  \item A study of social distancing measurement on public transportation based on the existing, low-resolution onboard surveillance cameras in the vehicles.
  \item We show that domain knowledge based on expected passenger behaviour on public transportation can improve the results by certain condition.
  \item We highlight the complexity of detecting passengers and social distancing automatically via object detection algorithms on real world CCTV footage and suggest how this could be improved.
\end{itemize}

\section{Related Work}
\label{sec:related-work}


Recently, taking advantage of the performance of CNN in detecting objects and people, there has been great interest in developing methods for the detection of social distancing which we review in this section.

For example, YOLO v3 has been used as the object detection architecture for detecting pedestrians in~\cite{kumari2021deep, magoo2021deep, shalini2021social, punn2020monitoring, mercaldo2021proposal, hou2020social}, primarily because it works better in a real-time setting~\cite{punn2020monitoring}. Other versions of YOLO architecture have also been used for object detection, such as YOLO v2~\cite{saponara2021implementing} and YOLO v4~\cite{rahim2021monitoring}. Other work in the literature has focused on comparing different object detection algorithms at detecting pedestrians, for example, the authors in~\cite{yang2021vision} compare Faster R-CNN and YOLO v4, the authors in~\cite{punn2020monitoring} compare Faster R-CNN, Single Shot Detector (SSD) and YOLO v3 and the authors in~\cite{zuo2021reference} used YOLO v3 after comparing to RetinaNet and Mask R-CNN.


In order to measure the distance between people, the authors in~\cite{kumari2021deep, magoo2021deep, yang2021vision} transformed the camera view into a bird's eye view to get the location of the detected person from the top-down perspective. The centroid of the detected person is used to measure the Euclidean distance to any other person in the image. The Euclidean distance is also used in \cite{pooranam2021safety,shalini2021social, mercaldo2021proposal, rahim2021monitoring, su2021novel} as well without any camera transformation for social distance measurement. In \cite{hou2020social}, the position of the detected pedestrian is estimated based on the bottom-centre point of the bounding box, rather than the centroid, before computing the distance between every pedestrian pair. Instead of tracking individual location, the work in \cite{zuo2021reference} measures the pedestrian density in a given image, via the use of a heatmap and temporal density distribution.

The author in~\cite{rezaei2020deepsocial} proposed a model, entitled DeepSOCIAL, based on a modified YOLO v3 to detect, track and estimate the distance between people. They applied the Simple Online and Real-time (SORT) tracking technique that uses Kalman filter and Hungarian optimisation to track and assign a unique ID to each detected people. To then detect the distance between people, they used inverse perspective geometric mapping.


The authors in \cite{varghese2021multimodal} proposed a multimodal Convolutional Variational Autoencoder (CVAE) fusion framework where the input data is from a combination of visual, infrared, audio and accelerometer sensors. The authors argue this multi-modality allows for a greater performance level than offered by considering only visuals.

Most of the work in social distancing measurement has been trained and evaluated on CCTV footage of pedestrians walking for example, from Oxford Town Centre~\cite{yang2021vision, punn2020monitoring, rezaei2020deepsocial} and New York~\cite{zuo2021reference}. Other studies have used data from low light environments\cite{rahim2021monitoring} and three different videos from town centers and beaches~\cite{mercaldo2021proposal}.

While earlier work focuses on CCTV footage that captures people mostly walking in a wide spacious area, in this paper we address the complexity of social distancing measurement on public transportation, especially those due to low-resolution images, non-optimal camera placement and physical limitations.  

\begin{figure}[!h]
    \centering
    \subfloat[V01]{\includegraphics[width=0.25\textwidth]{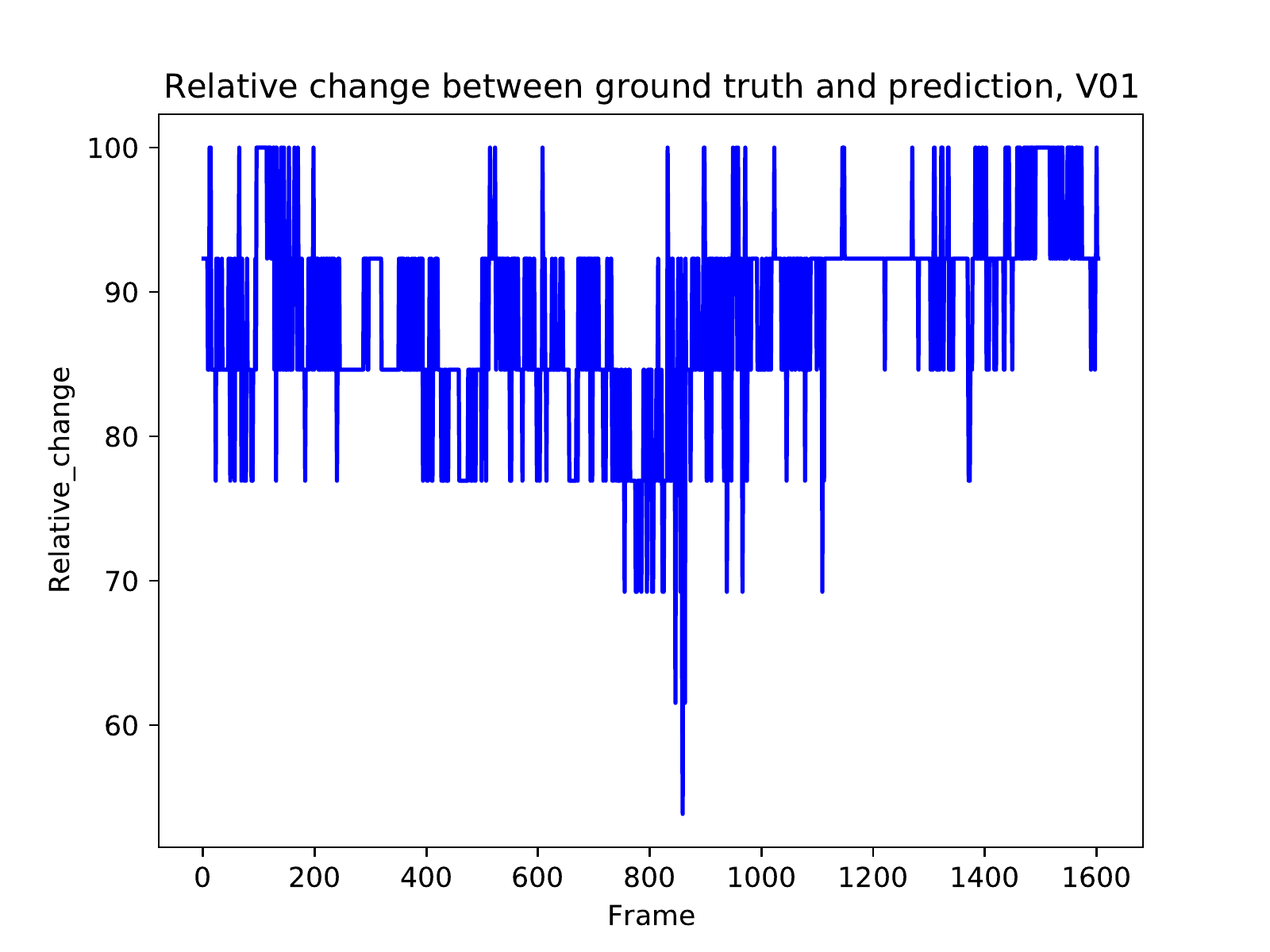}%
    \label{RC_V01}}
    \subfloat[V02]{\includegraphics[width=0.25\textwidth]{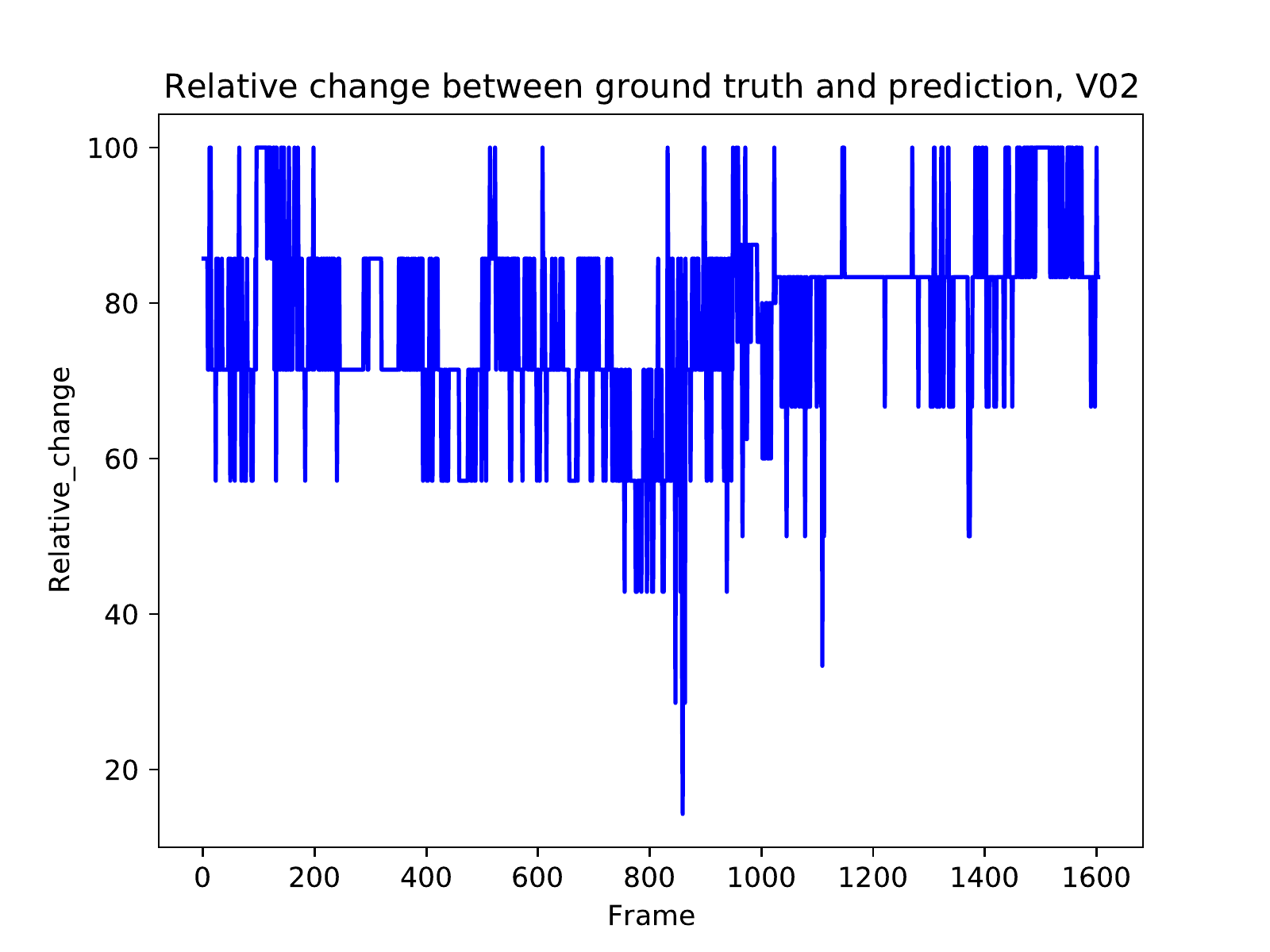}%
    \label{RC_V02}}
    \hfil
    \vskip -5pt
    \subfloat[V03]{\includegraphics[width=0.25\textwidth]{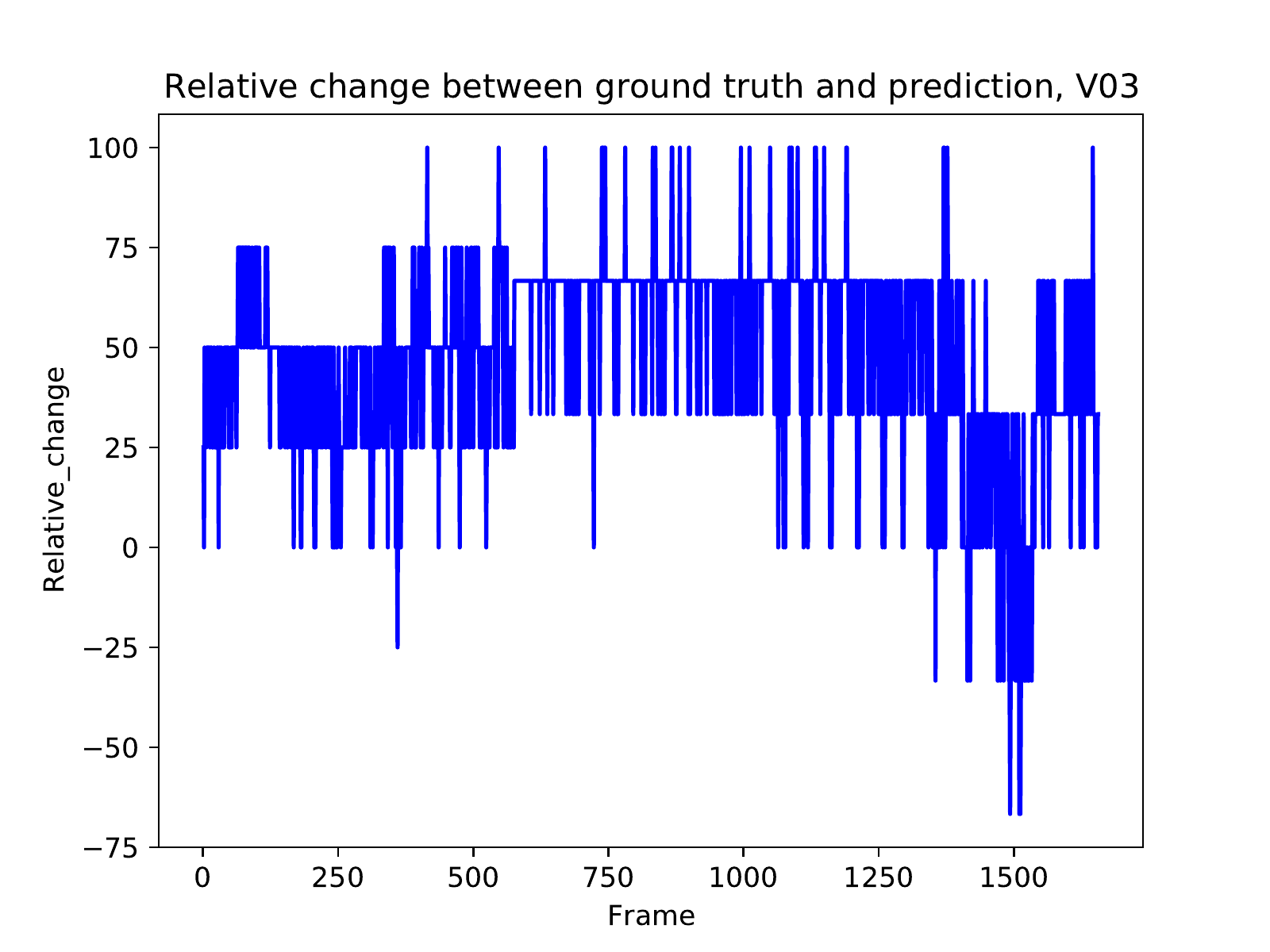}%
    \label{RC_V03}}
    \subfloat[V04]{\includegraphics[width=0.25\textwidth]{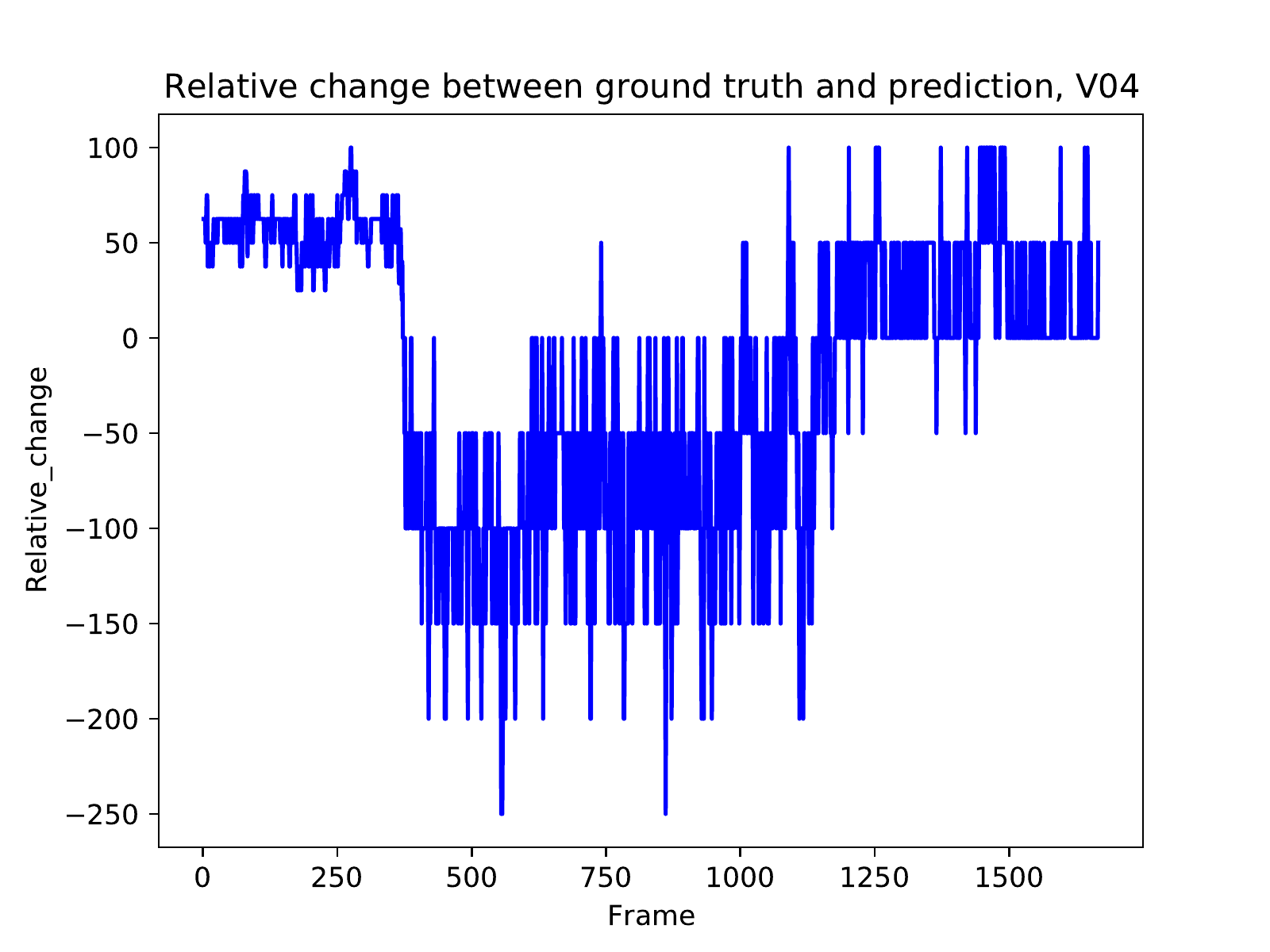}%
    \label{RC_V04}}
    \hfil
    \subfloat[V05]{\includegraphics[width=0.25\textwidth]{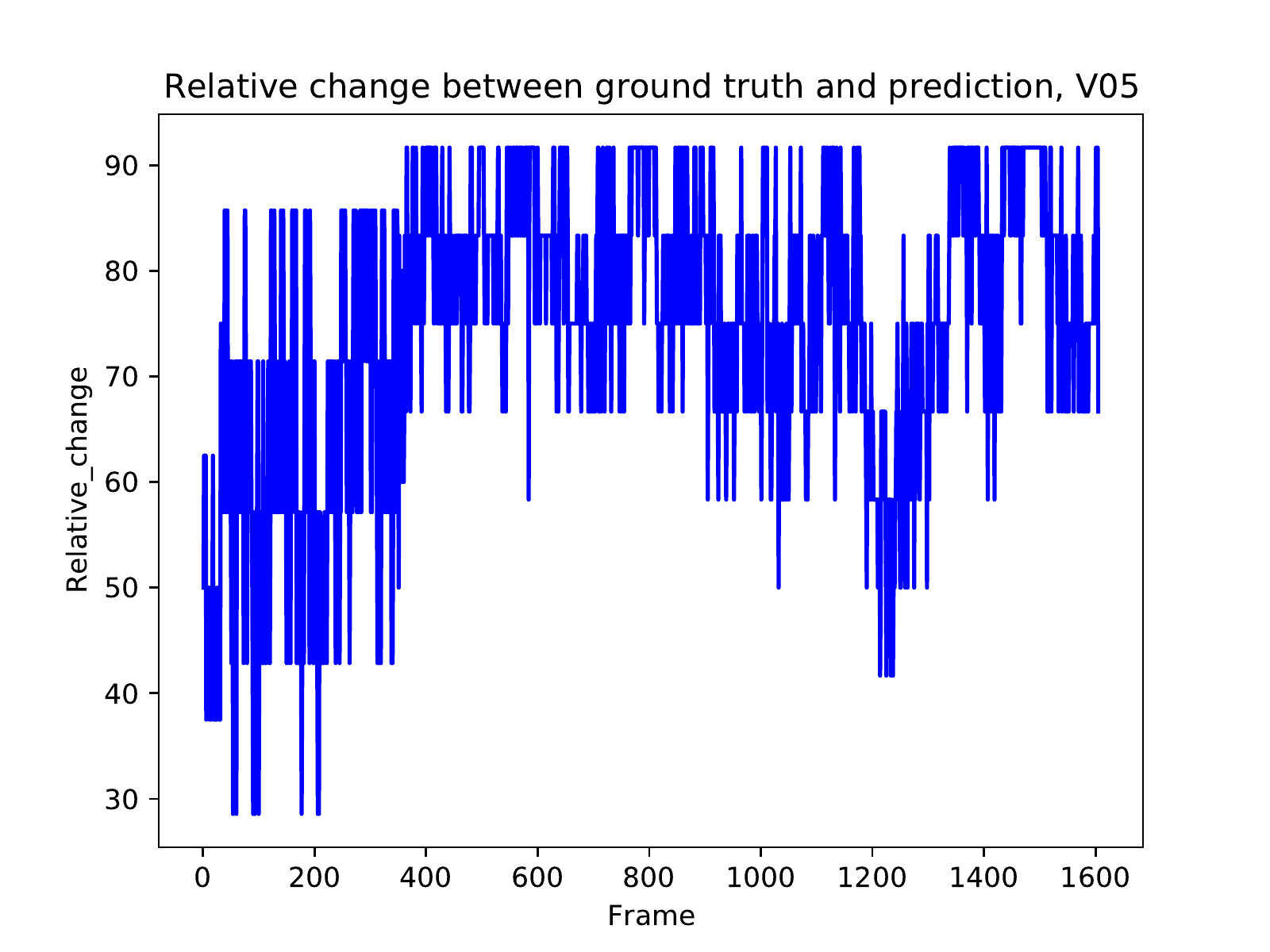}%
    \label{RC_V05}}
    \subfloat[V06]{\includegraphics[width=0.25\textwidth]{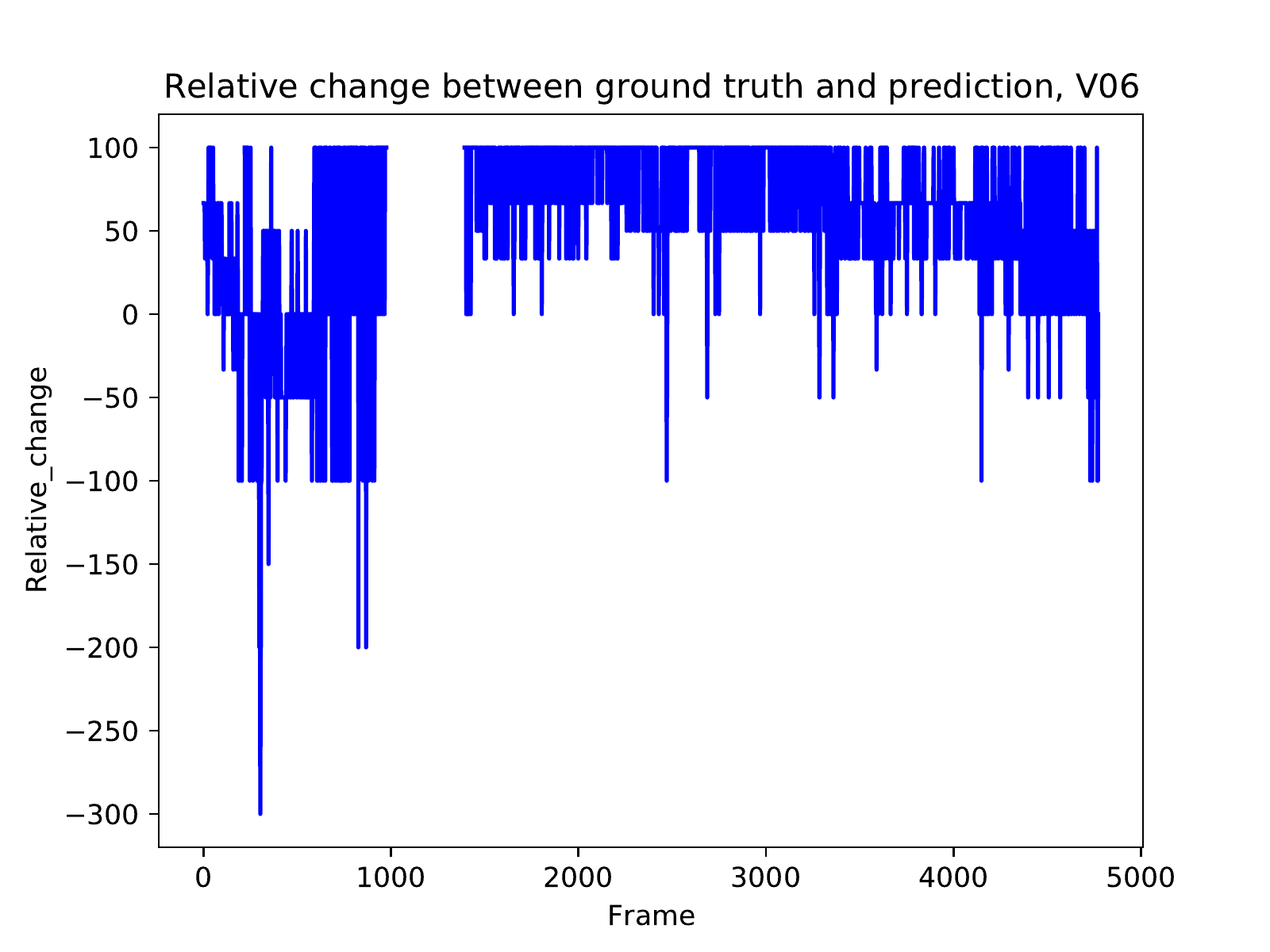}%
    \label{RC_V06}}
    \hfil
    \subfloat[V07]{\includegraphics[width=0.25\textwidth]{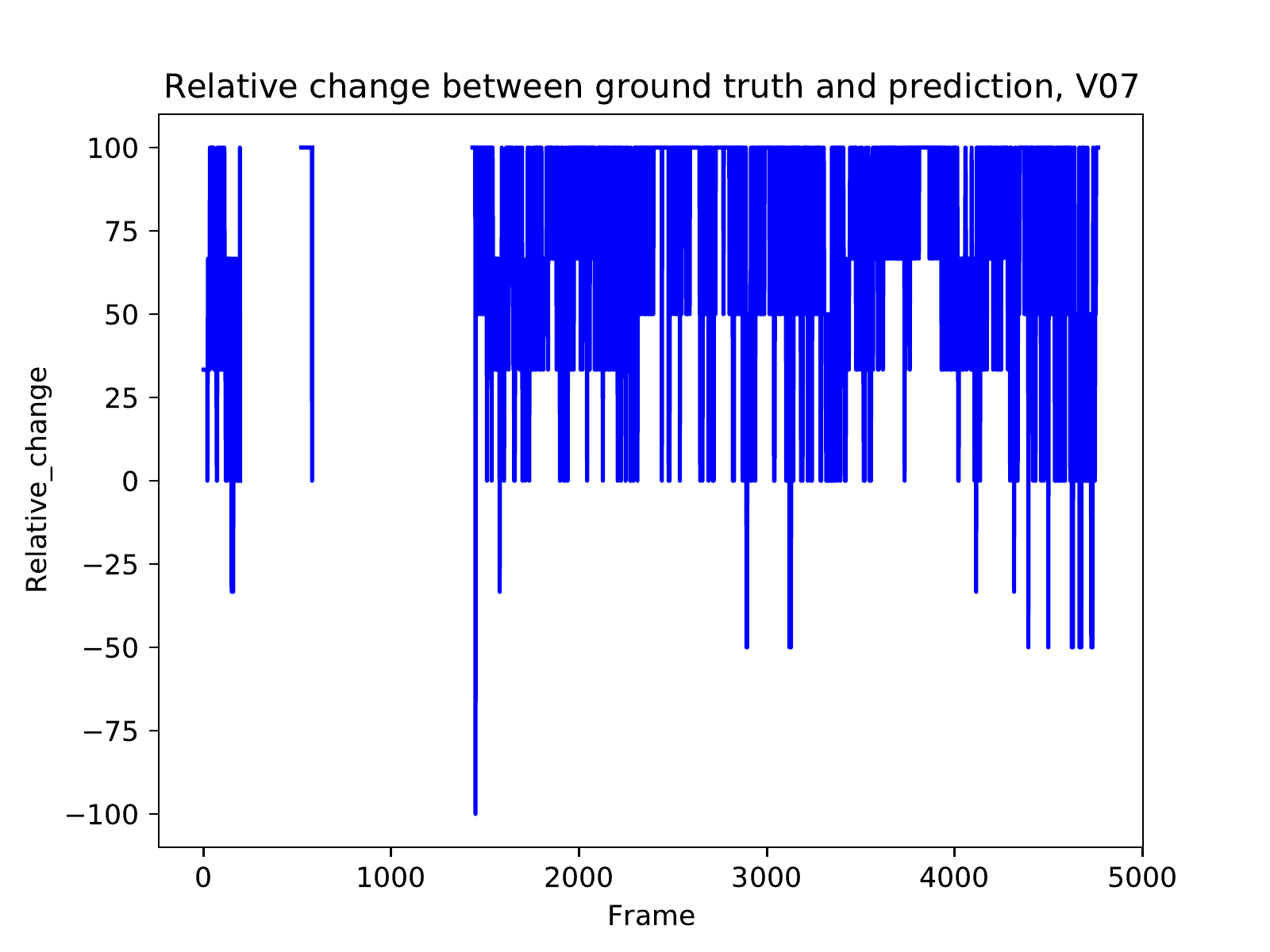}%
    \label{RC_V07}}
    \subfloat[V08]{\includegraphics[width=0.25\textwidth]{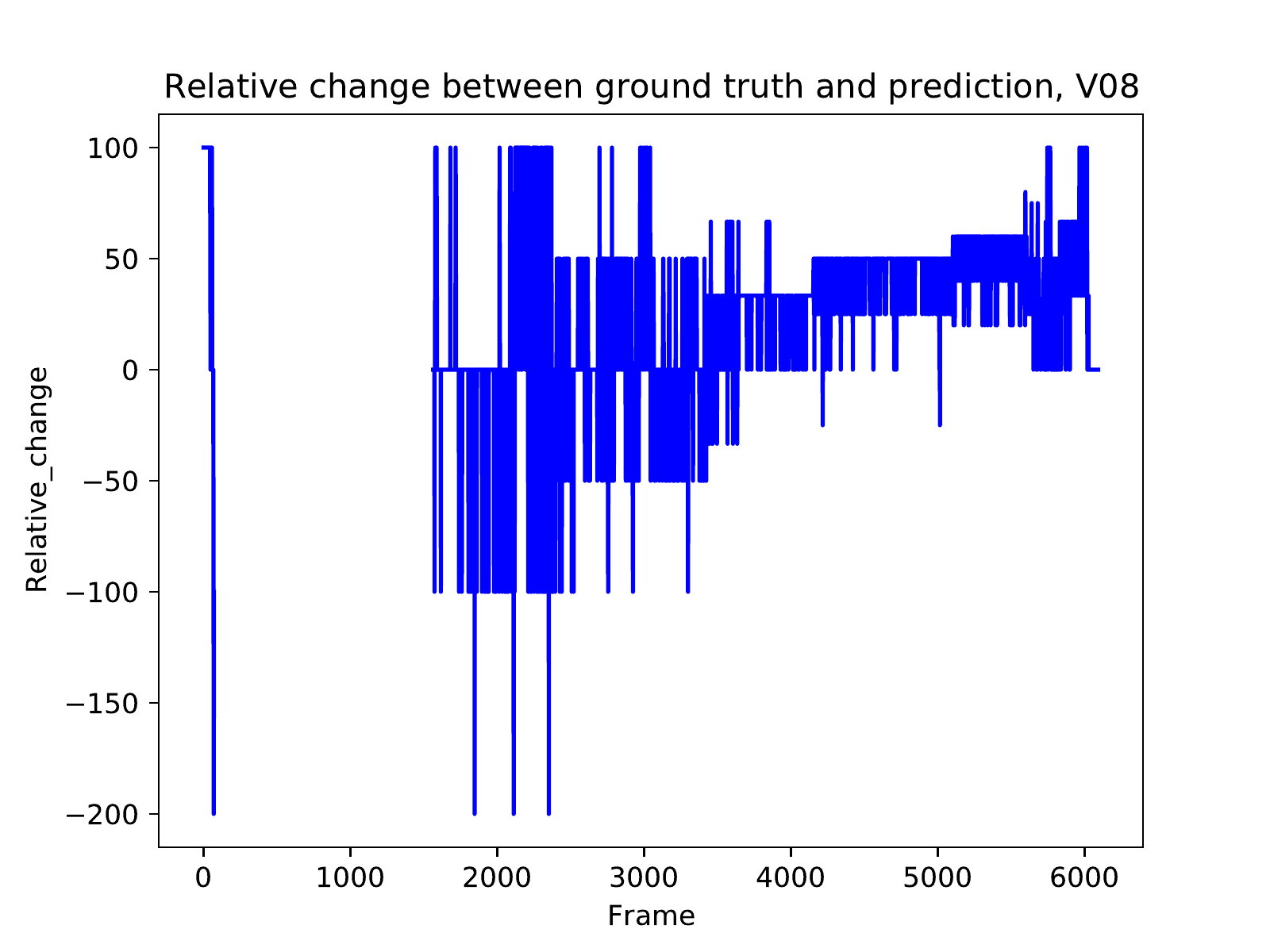}%
    \label{RC_V08}}
    \hfil
    \subfloat[V09]{\includegraphics[width=0.25\textwidth]{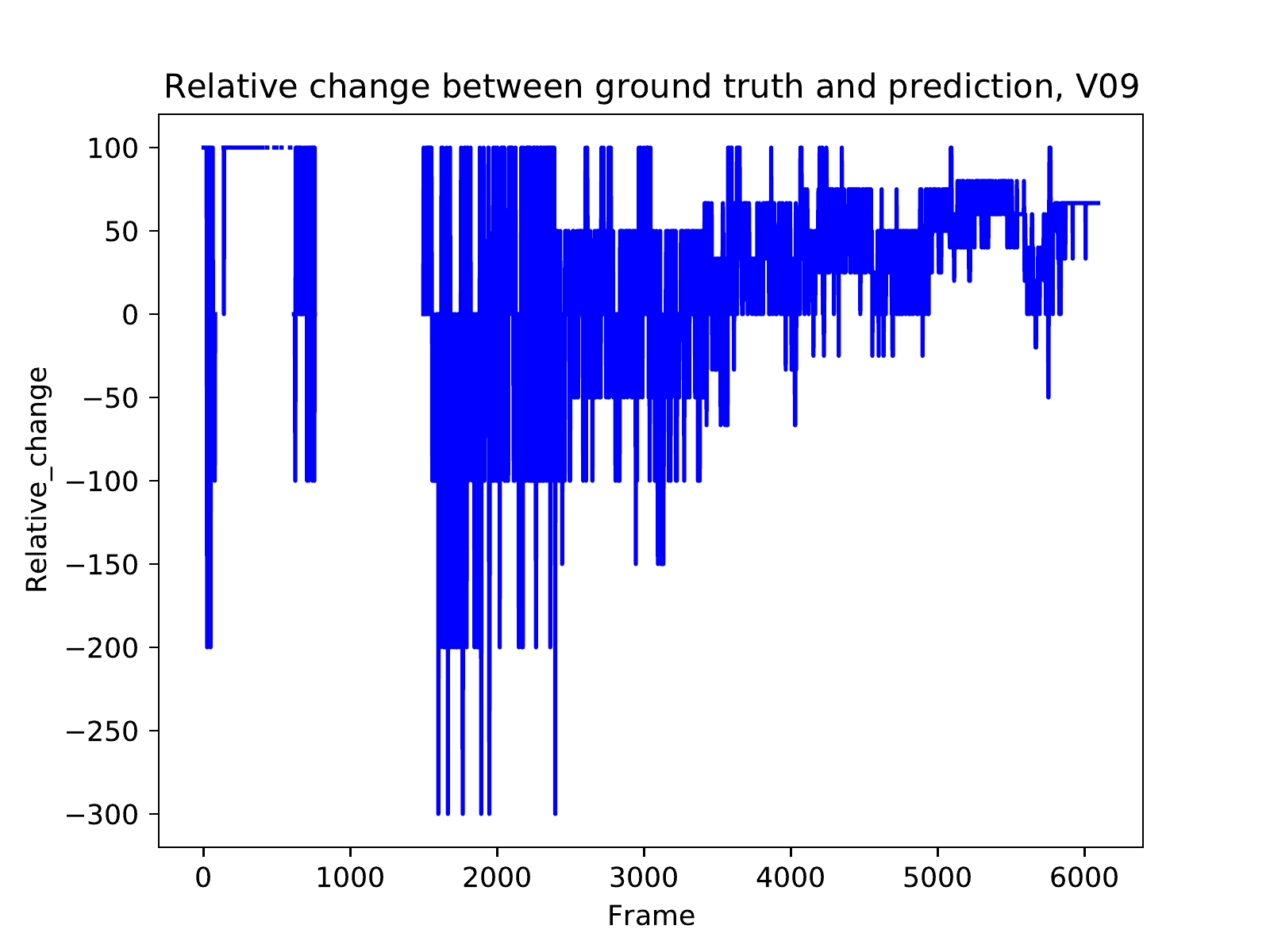}%
    \label{RC_V09}}
    \subfloat[V10]{\includegraphics[width=0.25\textwidth]{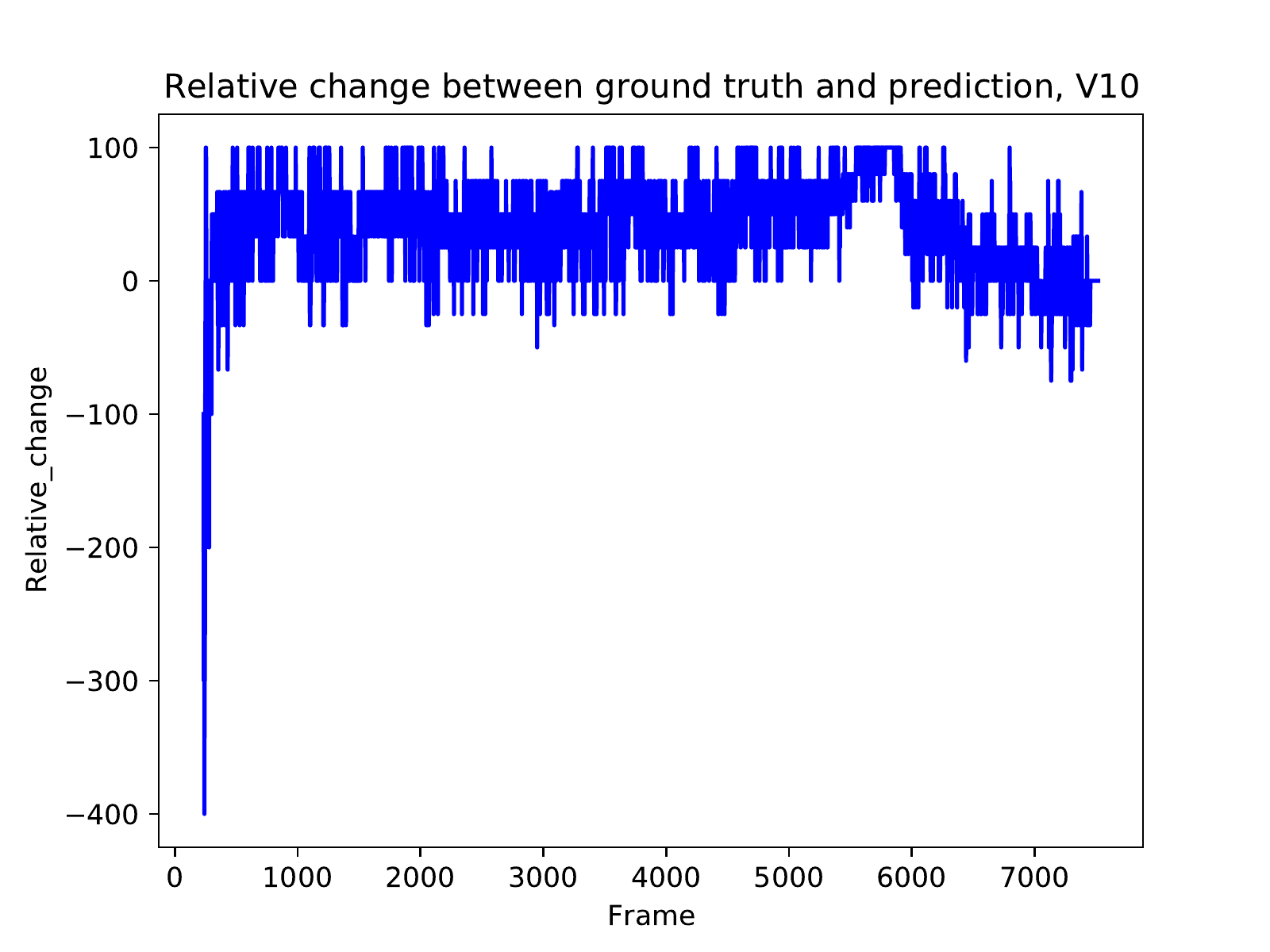}%
    \label{RC_V10}}
    \caption{Relative change between the headcount of ground truth and prediction.}
    \label{fig:RC}
  \end{figure} 
  
  \begin{figure*}[!t]
    \centering
    \subfloat[V01]{\includegraphics[width=0.2\textwidth]{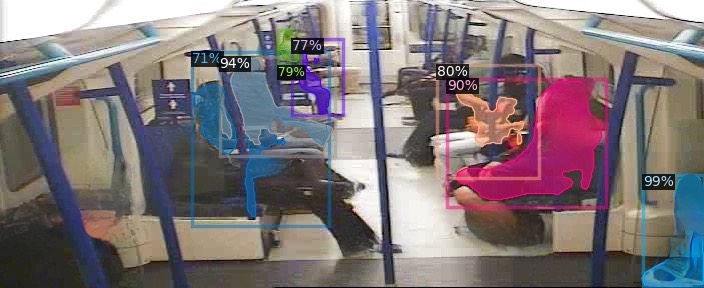}%
    \label{V01}}
    \subfloat[V02]{\includegraphics[width=0.2\textwidth]{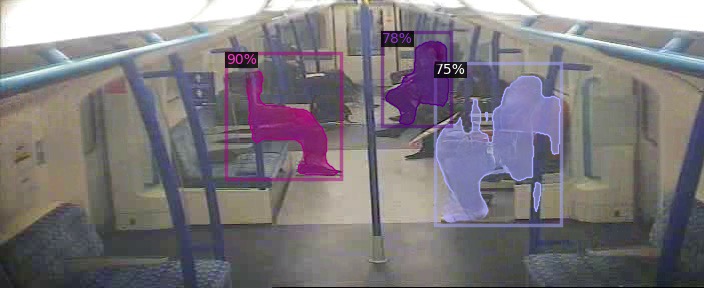}%
    \label{V02}}
    \subfloat[V03]{\includegraphics[width=0.2\textwidth]{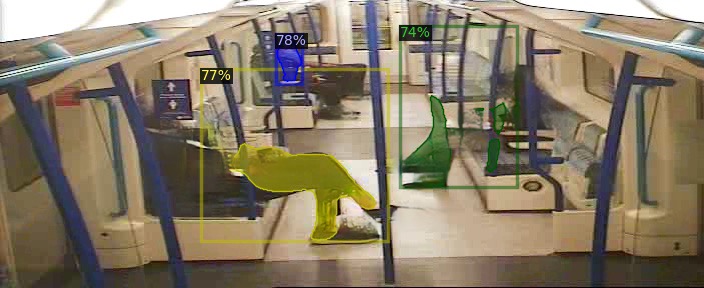}%
    \label{V03}}
    \hfil
    \subfloat[V04]{\includegraphics[width=0.2\textwidth]{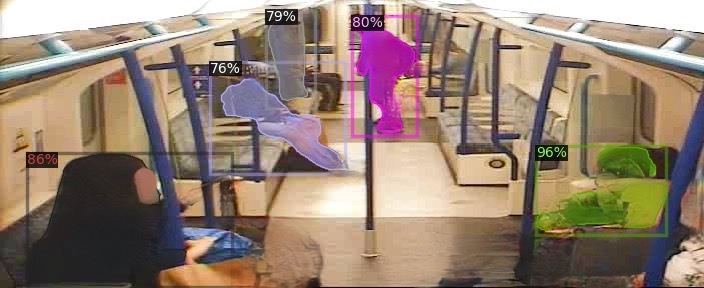}%
    \label{V04}}
    \subfloat[V05]{\includegraphics[width=0.2\textwidth]{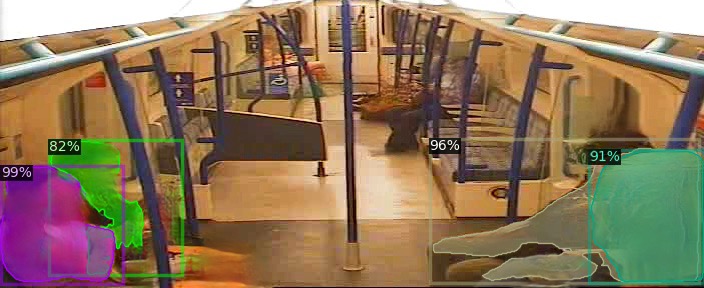}%
    \label{V05}}
    \hfil
    \subfloat[V06]{\includegraphics[width=0.2\textwidth]{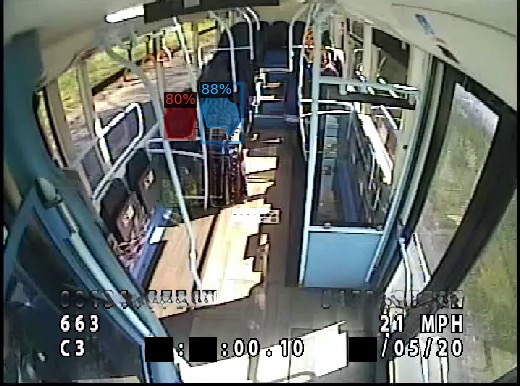}%
    \label{V06}}
    \subfloat[V07]{\includegraphics[width=0.2\textwidth]{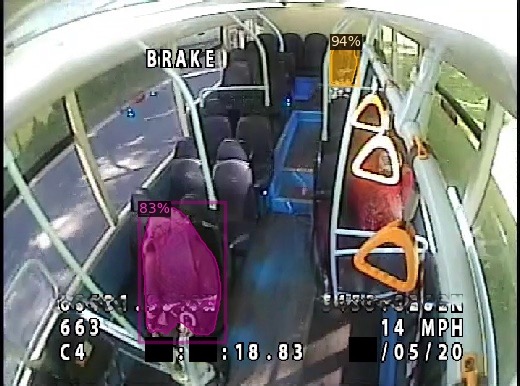}%
    \label{V07}}
    \subfloat[V08]{\includegraphics[width=0.2\textwidth]{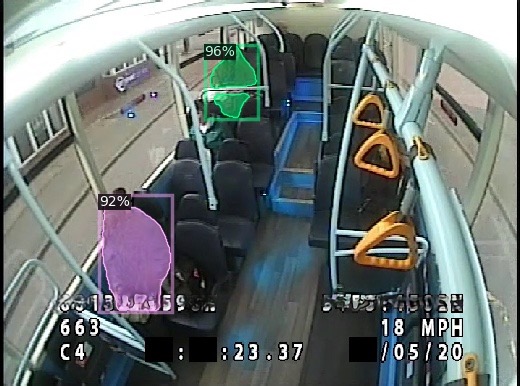}%
    \label{V08}}
    \subfloat[V09]{\includegraphics[width=0.2\textwidth]{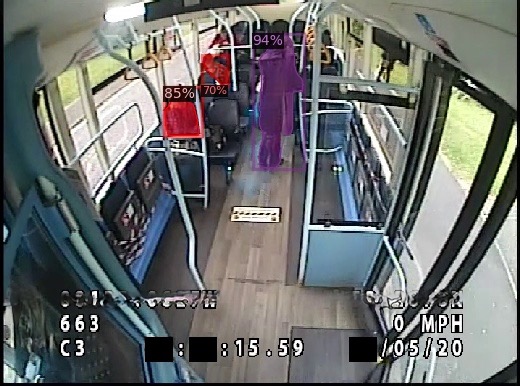}%
    \label{V09}}
    \subfloat[V10]{\includegraphics[width=0.2\textwidth]{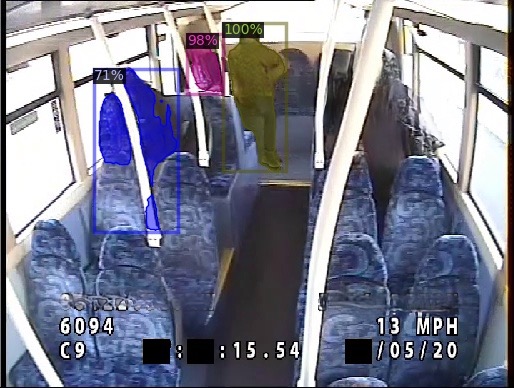}%
    \label{V10}}
    \caption{Passenger detection via Mask R-CNN}
    \label{fig:Headcount_example}
  \end{figure*}



\section{Dataset}
\label{sec:Dataset}

This paper analyses the complexity of applying the social distancing measurement via onboard low-resolution CCTV, especially public transportation. Our analysis is based on footage from a London underground line and from Go North East busses. 

The majority of the London Underground lines have high-resolution CCTV cameras. However, this study investigates the complexity of detecting  passengers not only in low-resolution images that can be found on a specific line on London underground and busses but also because of the occlusion caused by the camera's physical location. 

We randomly choose five clips from the Go North East bus journey and five from a London underground train to visualise the outcome of this analysis. For privacy reasons, all the images in this paper have been post-processed with a Gaussian blur after the experiments to hide any passengers and their belongings.

\section{Methodology}
\label{sec:methods}

This section discusses three different state-of-the-art object detection approaches on images from low-resolution onboard surveillance cameras on public transportation to assess how well the pre-trained networks detect the passengers. We also present a post-processing technique we used to investigate if we can improve the quality of the detections.

\subsection{Object Detection}
\label{sec:objdetection}

Object Detection is one of the most used applications within Computer Vision and has been widely used in object tracking, autonomous vehicle, medical imaging and many more application. The goal of object detection is to locate each object in an image \cite{Vahab2019applications}. In this paper, we show three different object detection algorithms to see if the algorithms can detect the passengers using the onboard CCTV camera footage from the bus and underground train.

\subsubsection{Faster R-CNN}
\label{sec:frcnn}

The Faster R-CNN is a region-based CNN that introduces another convolutional network (on top of the CNN for the classification task) as Region Proposal Network (RPN) to generate the pixel coordinates of the potential objects in an image \cite{ren2015faster}. A region-based network is a network that has a region proposal module that generates a variety of bounding boxes with different sizes and aspect ratios as the proposed object regions. This network will return object classes with the score and the bounding boxes for each object as the pixel coordinates \cite{girshick2014rich}.

\subsubsection{Mask R-CNN}
\label{sec:mrcnn}

Mask R-CNN used the RPN from Faster R-CNN, but instead of just the bounding box, it indicates the binary mask of each object and the bounding box. The binary mask gives 1 to the pixel that belongs to the objects and 0 otherwise. Mask R-CNN is considered an instance segmentation because of the mask segmentation, given the pixel by pixel location and the bounding box of every object. The granularity can be beneficial in applications such as autonomous vehicles and pose estimation, where precision is essential \cite{he2017mask}.






      

      



          
          
        
      
      
      
    


  
  

\begin{figure}[!h]
  \centering
  \subfloat[Passenger sitting down.]{\includegraphics[width=0.32\textwidth]{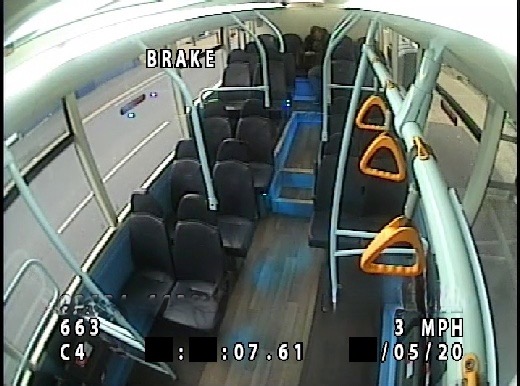}%
  \label{S00}}
  \hfil
  \subfloat[Passenger standing up.]{\includegraphics[width=0.32\textwidth]{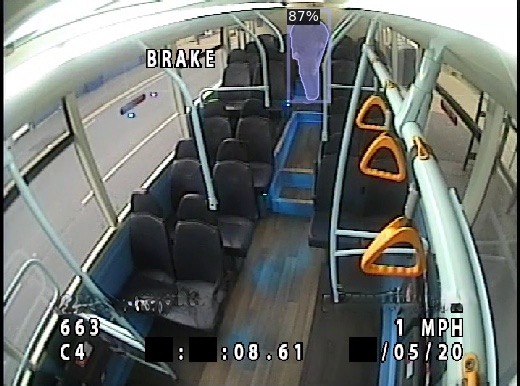}%
  \label{S01}}

  \caption{Comparison of the Mask R-CNN in detecting passenger standing up as compared to the same passenger sitting down.}
  \vskip -5pt
  \label{fig:tand}
\end{figure}

\subsubsection{YOLOv3}
\label{sec:yolov3}

Unlike region-based object detection CNN use two networks, one for object localisation and another for feature extraction, You Only Look Once (YOLO) introduces single-stage architecture using the regression-based method \cite{redmon2016you}. YOLO v3 has a deep end-to-end CNN that deals with feature extraction and object localisation, making this unified architecture very quick to train \cite{redmon2018yolov3}.

We show examples of these three object detection approaches as in Figure \ref{fig:Modelcompare}. As the performance of Mask R-CNN is better than Faster R-CNN or YOLOv3, in this paper, we use Mask R-CNN running on Detectron2 framework \cite{wu2019detectron2} as our passenger detection.

\subsection{Post Processing}
\label{sec:postprocess}

There are some possible false positives in the footage that caused by the degradation of images or the camera angle, especially on the bus that captures partial images from outside the vehicle. In this paper, we explore the way to eliminate any false positive detection by using bounding box analysis. We are interested in reducing the false positives via the usage of domain knowledge on the expected behaviour of the passengers on public transportation by eliminating any bounding boxes from the prediction that only appear for a few frames.

The assumption is that, once people are on the bus, their bounding boxes will frequently appear within the same region for at least a few minutes. If a ``person" is detected for less than a certain number of frames, we can assume it is a misidentification, for example, a pedestrian or cyclist outside of the bus.

Our bounding box analysis is based on a few assumptions:

\begin{itemize}
  
  \item Most of the time, all passengers are roughly in the same place (they are sitting in the same seat) throughout the journey.
  \item There is no more than one false positive in a frame.
  \item We vary the frame window between 4 frames to 300 frames (1 second to 60 seconds given 4fps of the video).
  \item We vary the centre of the bounding box within $\pm$10 and $\pm$15 pixels.
  
\end{itemize}

\section{Results and Discussion}
\label{sec:results}

\subsection{Passenger Detection}
\label{sec:results_detection}

Based on the preliminary experiments on comparing three different object detection models, we use Mask R-CNN running on the Detectron2 framework for passenger detection in this paper. We compare the headcount from the Mask R-CNN to the ground truth to see how accurately the model detects passengers on public transportation. We randomly choose five different camera footages from the bus journey and five from the underground train, and manually count the number of passengers of each journey as the ground truth headcount. We also randomly select a frame range where the headcount is constant to monitor the performance of the Mask R-CNN.

To study whether the model under or over predicts, we plot the relative change in percentage as in Figure~\ref{fig:RC}. We measure the relative changes of the headcount for every frame based on $:$

\begin{equation}
  RC=\frac{y_{i}-x_{i}}{y_{i}},
  \label{eq:rc}
\end{equation} 

where $y_i$ is the ground truth number and $x_i$ is the prediction number. The percentage above zero indicates under prediction, while the percentage less than zero indicates over prediction (False Positive).

Table~\ref{table:headcount} shows the headcount of the prediction using Mask R-CNN to the ground truth on the frames where the ground truth headcounts are constant (i.e. between stops) to show the performance of Mask R-CNN on detecting passengers. We also calculate the mean absolute relative changes to see average relative changes across all frames on the footage. 

Figure~\ref{fig:RC} and Table~\ref{table:headcount} show that the predictions often differ greatly from the ground truth headcount. The bus footage plots indicate that false positives happen in all the footage. Some of the examples of object detection via Mask R-CNN can be seen in Figure~\ref{fig:Headcount_example}.

The current literature in Section~\ref{sec:related-work} focuses on pedestrian detection, where the pedestrian in the footage is mostly standing up, which shows more visible features than people sitting down. Figure~\ref{fig:tand} shows an example of a passenger being easily detected when standing up compared to sitting down.

\begin{figure}[!h]
  \centering
  \subfloat[A passenger is blocking another passenger.]{\includegraphics[width=0.32\textwidth]{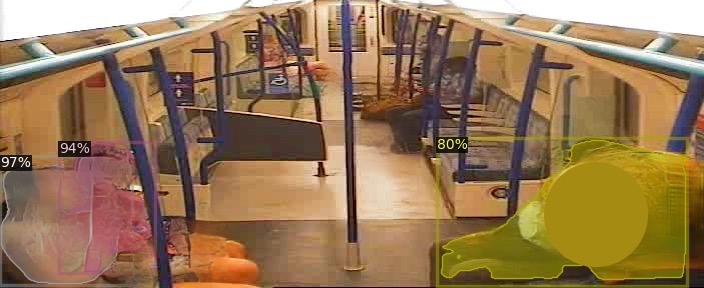}%
  \label{Bad00}}
  \hfill
  \subfloat[Two passengers have been detected as an object.]{\includegraphics[width=0.32\textwidth]{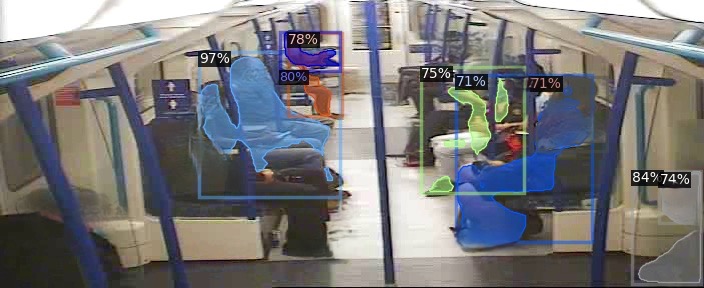}%
  \label{Bad01}}
  \hfill
  \subfloat[Pedestrian on the street.]{\includegraphics[width=0.32\textwidth]{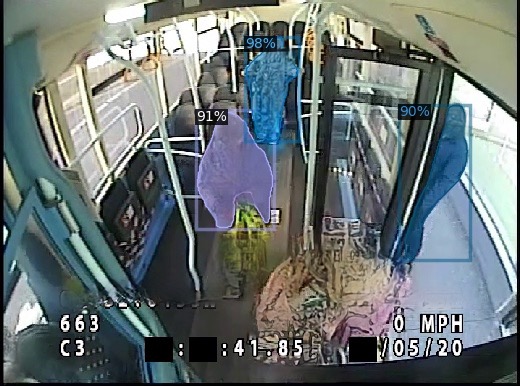}%
  \label{Bad02}}
  \hfill
  \subfloat[Cyclist on the street.]{\includegraphics[width=0.32\textwidth]{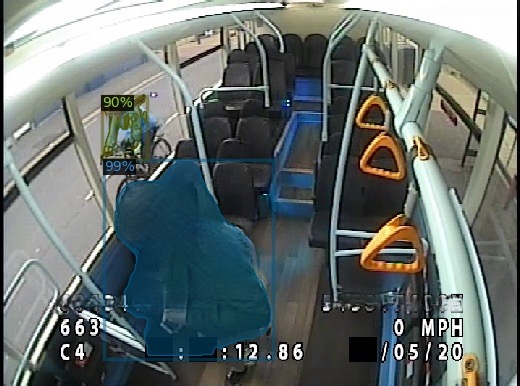}%
  \label{Bad03}}

  \caption{Example of the incorrect detection on the public transportation footages.}
  \label{fig:Bad}
\end{figure}

Object detection works very well when the model can learn features that discriminate between different classes:- in our case, between a person and a non-person. However, we have to work with low-resolution CCTV images, making it hard for the object detection network (Mask R-CNN) to identify the features within the images. Apart from that, there are many passengers that are occluded because of the CCTV's physical location, for example, in Figure~\ref{fig:Bad}.

\begin{figure*}[!h]
  \centering
  \subfloat[V06]{\includegraphics[width=0.2\textwidth]{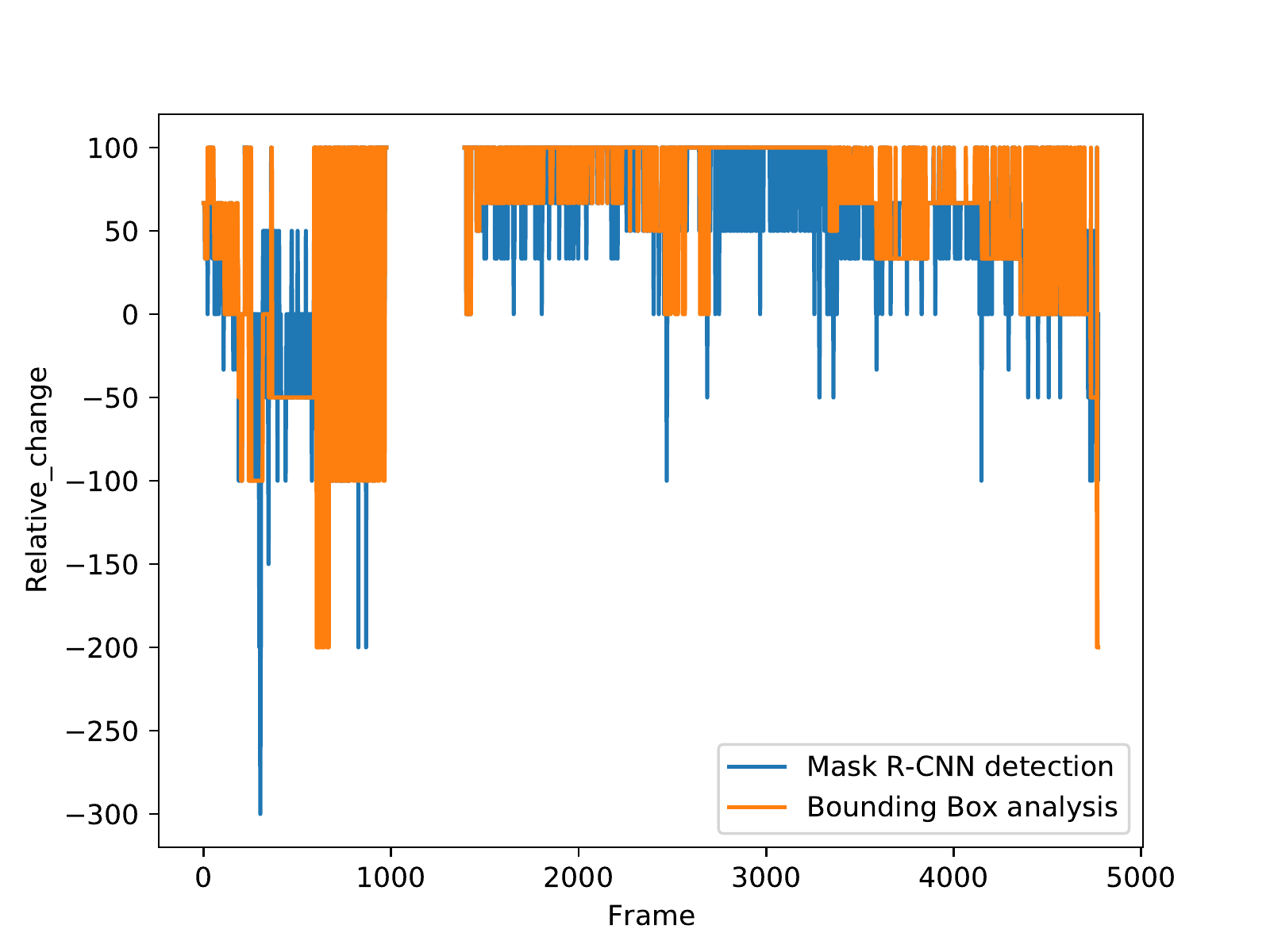}%
  \label{RCBB_V06}}
  \subfloat[V07]{\includegraphics[width=0.2\textwidth]{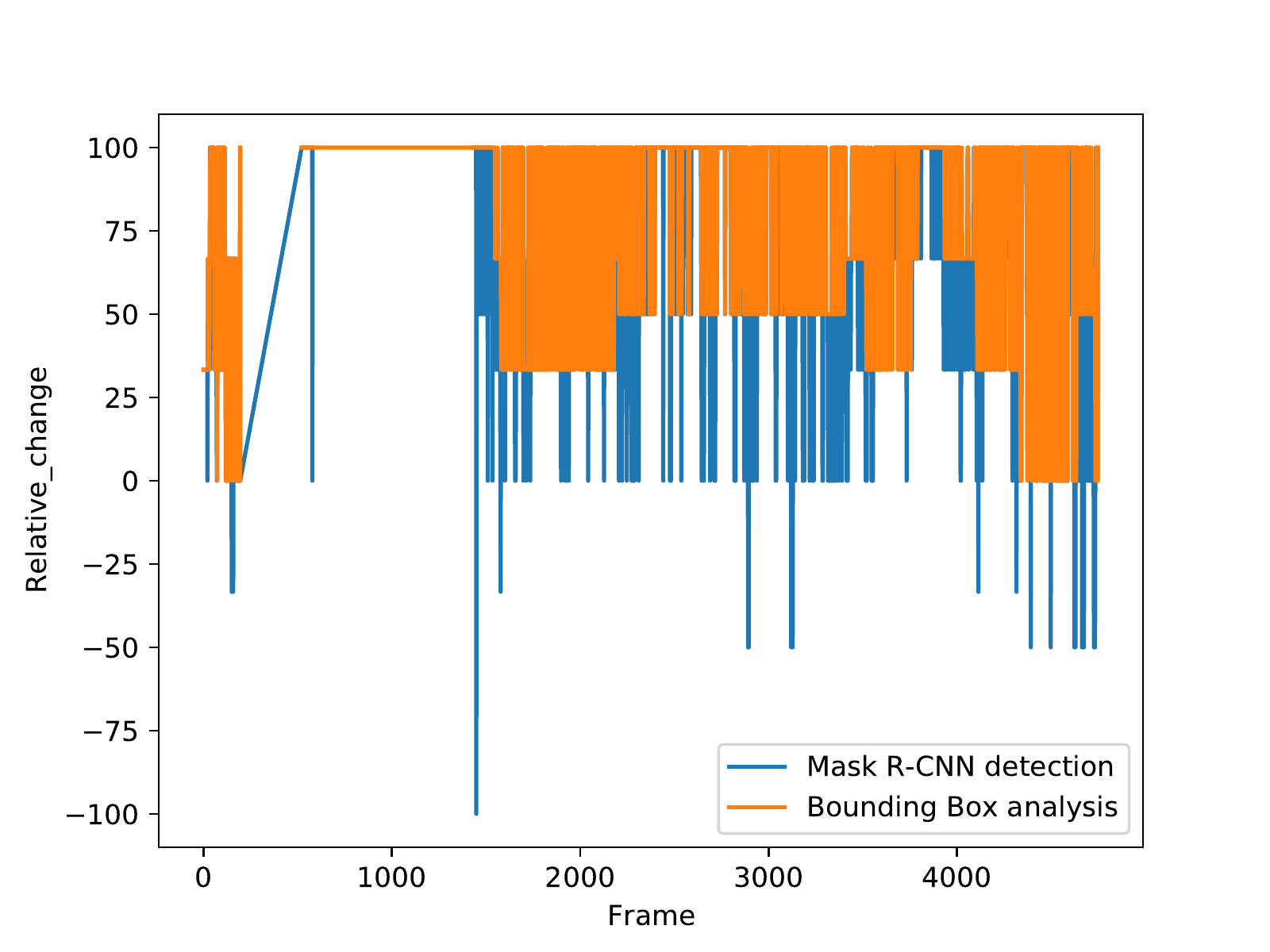}%
  \label{RCBB_V07}}
  \subfloat[V08]{\includegraphics[width=0.2\textwidth]{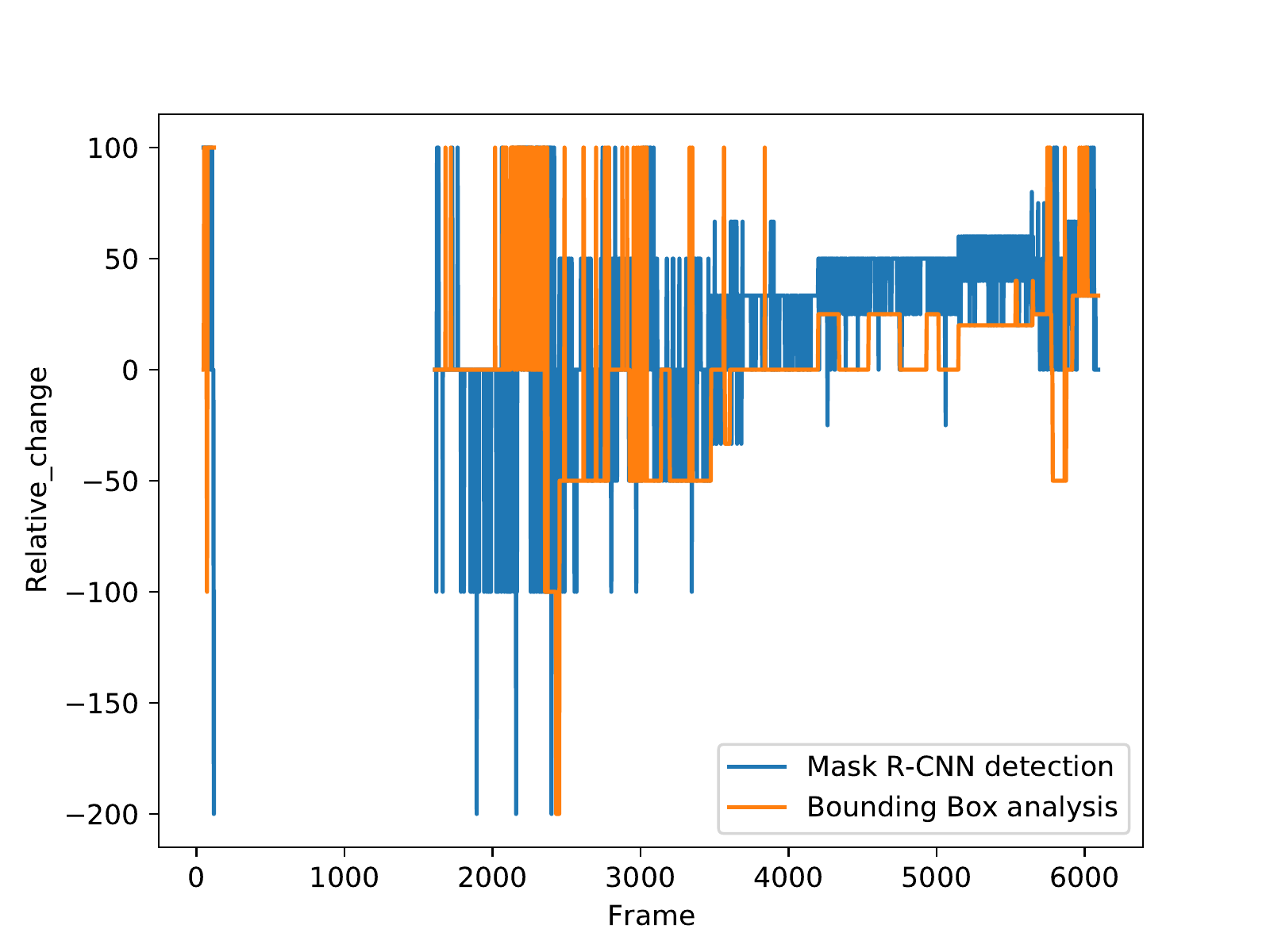}%
  \label{RCBB_V08}}
  \subfloat[V09]{\includegraphics[width=0.2\textwidth]{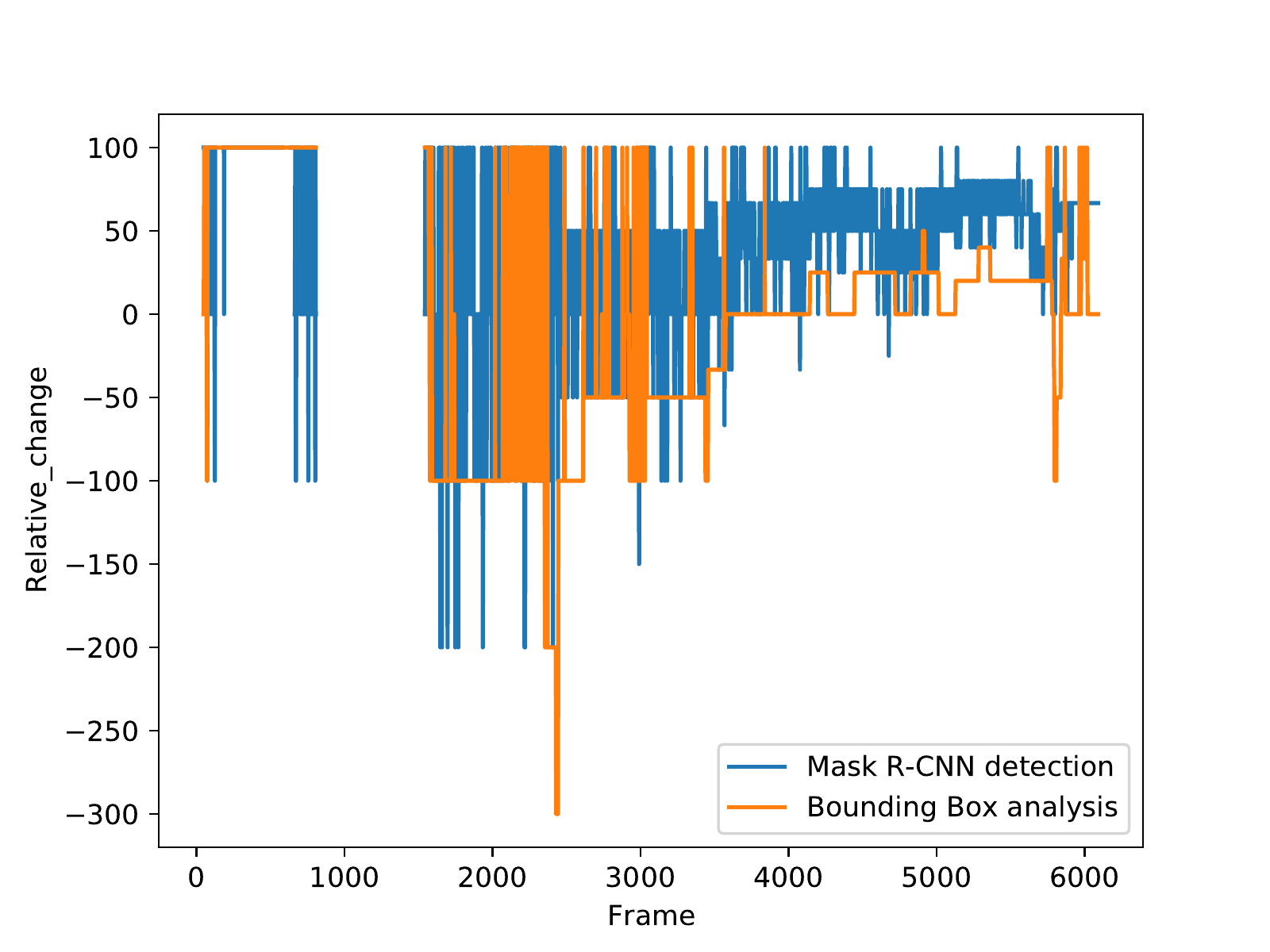}%
  \label{RCBB_V09}}
  \subfloat[V10]{\includegraphics[width=0.2\textwidth]{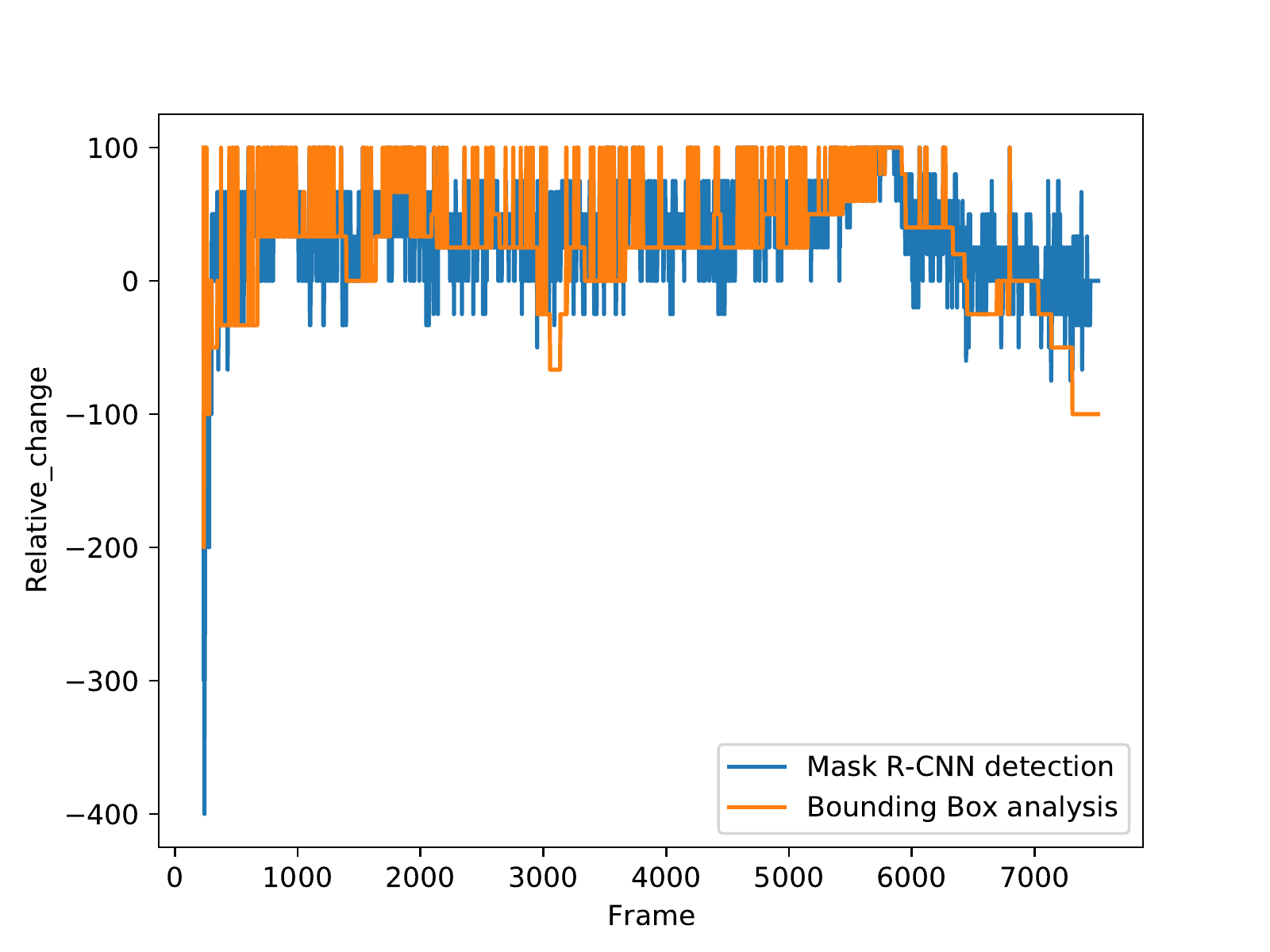}%
  \label{RCBB_V10}}
  \caption{Relative change between the headcount of ground truth and prediction before and after the bounding box analysis.}
  \vskip -10pt
  \label{fig:BB_RC}
\end{figure*}

\begin{figure*}[!h]
  \centering
  \subfloat[V01]{\includegraphics[width=0.2\textwidth]{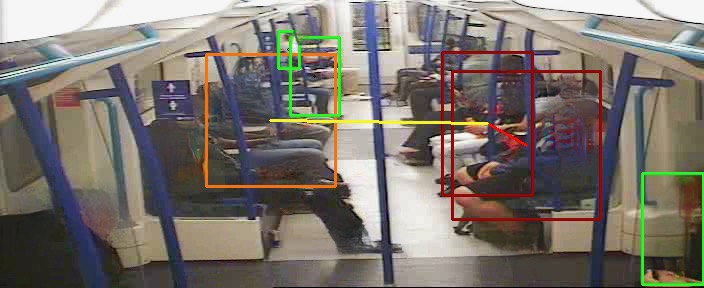}%
  \label{V01}}
  \subfloat[V02]{\includegraphics[width=0.2\textwidth]{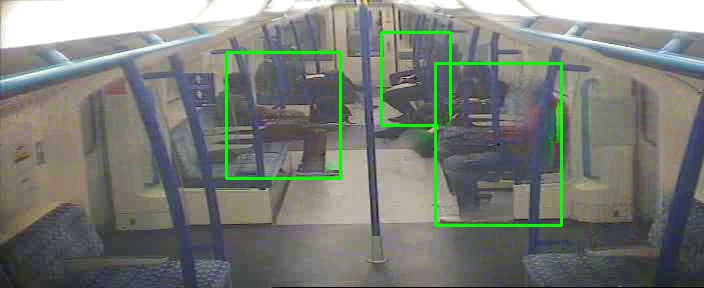}%
  \label{V02}}
  \subfloat[V03]{\includegraphics[width=0.2\textwidth]{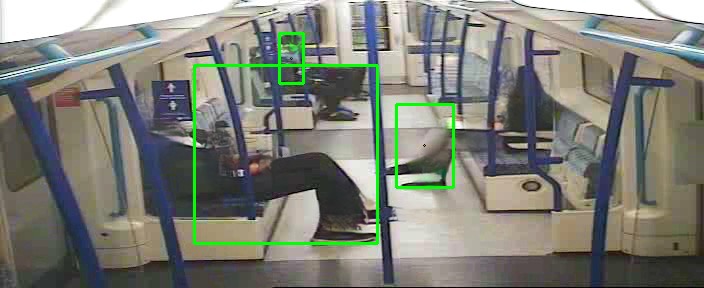}%
  \label{V03}}
  \subfloat[V04]{\includegraphics[width=0.2\textwidth]{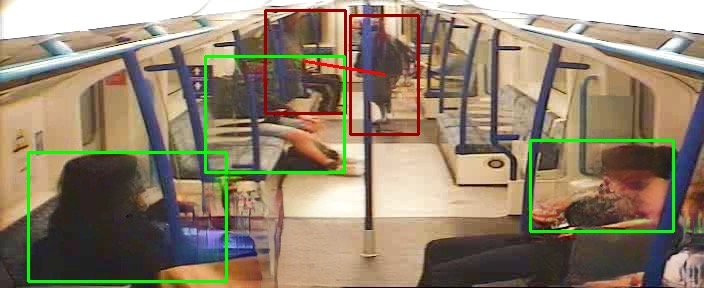}%
  \label{V04}}
  \subfloat[V05]{\includegraphics[width=0.2\textwidth]{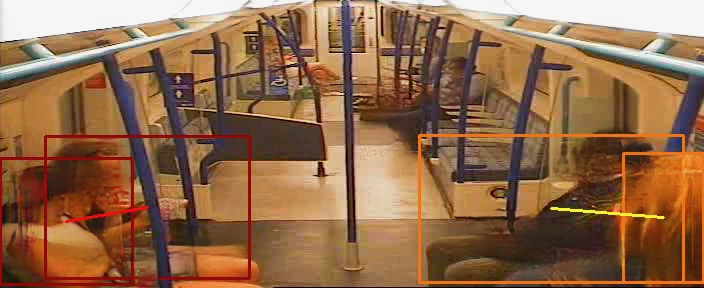}%
  \label{V05}}
  \hfil
  \subfloat[V06]{\includegraphics[width=0.2\textwidth]{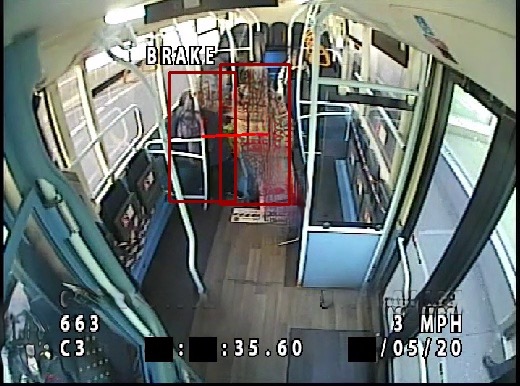}%
  \label{V06}}
  \subfloat[V07]{\includegraphics[width=0.2\textwidth]{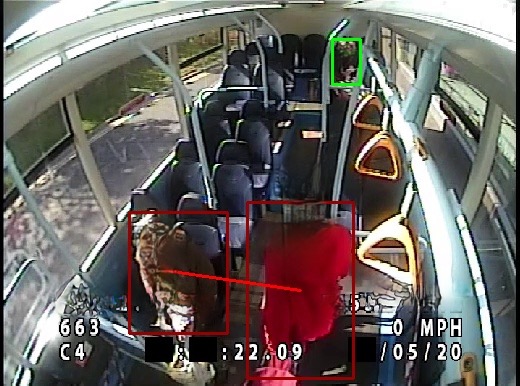}%
  \label{V07}}
  \subfloat[V08]{\includegraphics[width=0.2\textwidth]{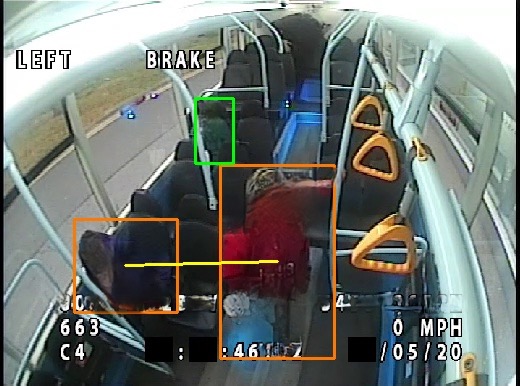}%
  \label{V08}}
  \subfloat[V09]{\includegraphics[width=0.2\textwidth]{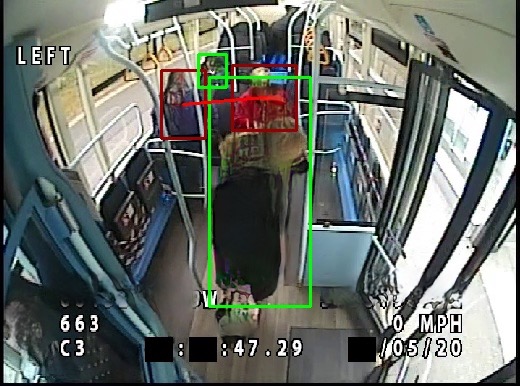}%
  \label{V09}}
  \subfloat[V10]{\includegraphics[width=0.2\textwidth]{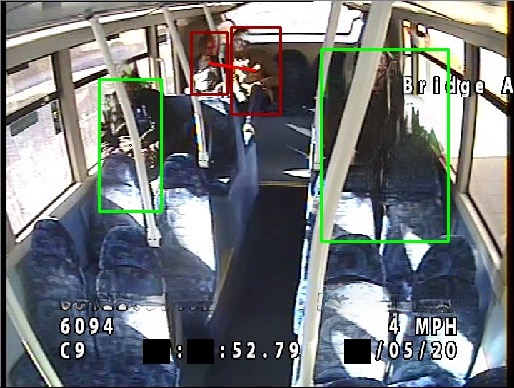}%
  \label{V10}}
  \caption{Social distancing measurement}
  \vskip -10pt
  \label{fig:SD_train}
\end{figure*}

In Figure \ref{Bad00}, we can see that a passenger is blocking another passenger, and the Mask R-CNN is blending this passenger's head to another passenger's lower body as a detection. Similarly, two passengers in Figure \ref{Bad01} have been detected as a passenger. There are pedestrians outside of the bus in Figure \ref{Bad02} and Figure \ref{Bad03} that have been detected, contributing to the detection's false positive. Additionally, a passenger in Figure \ref{Bad02} has not been detected, perhaps because of the unusual angle of the passenger and the hardcoded metadata causing some occlusion of the passenger.

\begin{table}[h!]
  \centering
  \caption{The number of passengers based on ground truth headcount and prediction headcount, and the mean absolute relative change (Mean Abs RC) between the headcounts.}
  \label{table:headcount}

  \begin{tabu}{c c c c}
  \hline\hline
  \textbf{Footage} & \textbf{Ground truth} & \textbf{Prediction} & \textbf{Mean Abs RC}\\
  \hline\hline
  \textbf{V01} & 13  & 7  & 60.5$\%$\T\\
  \textbf{V02} & 7  & 3 & 81.5$\%$ \\
  \textbf{V03} & 4  & 3 & 58.6$\%$\\
  \textbf{V04} & 8  & 5  & 55.1$\%$\\
  \textbf{V05} & 8  & 4  & 77.7$\%$\\
  \textbf{V06} & 3  & 2  & 73.3$\%$\\
  \textbf{V07} & 3  & 3  & 80.7$\%$\\
  \textbf{V08} & 3  & 2 & 36.0$\%$\\
  \textbf{V09} & 4  & 3 & 51.2$\%$\\
  \textbf{V10} & 5  & 4  & 53.8$\%$ \\
  \hline\hline
  \end{tabu}

\end{table}

In addition to the low-resolution camera, the physical location of the surveillance camera on the vehicle can also be an issue. For example;

\begin{itemize}
  
  \item Having a camera far away from the passenger seats' location results in some people only appearing as a few pixels in the frame.
  \item Other obstacles may occlude a clear line of sight to the passengers, such as seats, handrails, or other passengers in front of the others.
  \item As the CCTV also captures partial images from outside of the vehicle, false positives may occur, especially if any pedestrians or cyclists are captured within the frame.
  \item The inconsistent brightness on the vehicle will either create shadows that can be mistaken as a person or may result in low illumination in the images, again degrading the network's performance.

\end{itemize}

\subsection{Bounding Box Analysis}
\label{sec:results_bb}

Based on Figure \ref{fig:RC} in Section \ref{sec:results_detection}, all the bus footages (V06, V07, V08, V09, V10) have a significant number of false-positive detections. From the observation, the location of the CCTV cameras and the positions and the large windows contribute to the false positive detections on the bus. As the CCTV captures partial images from outside of the bus, false positives may occur, as when pedestrians or cyclists are captured within the frame. In addition, the inconsistent brightness on the bus will either create shadows that can be mistaken as a passenger, again degrading the performance of the object detection network. Therefore, we ran the bounding box analysis to investigate if our assumptions are correct and improve the relative change of passenger detection.

As shown in Figure \ref{fig:BB_RC}, this bounding box analysis method manages to remove some of the false positives. However, some of the true positives have been removed as well. The mean absolute relative change for these five videos now are; 76.1$\%$, 77.4$\%$, 23.2$\%$, 51.5$\%$, and 40.4$\%$. One of the observations that contribute to this is some of the passengers are sitting where they are obscured from a clear line of sight of the CCTV camera. Thus, the only time they were detected was when they walked to their seat or walked out from the bus. By eliminating the bounding box that appears less frequent around the same region, the detection of these passengers will have eliminated as well, resulting in them not being counted.




\begin{figure}[!h]
  \centering
  \subfloat[A part of the passenger has been detected.]{\includegraphics[width=0.35\textwidth]{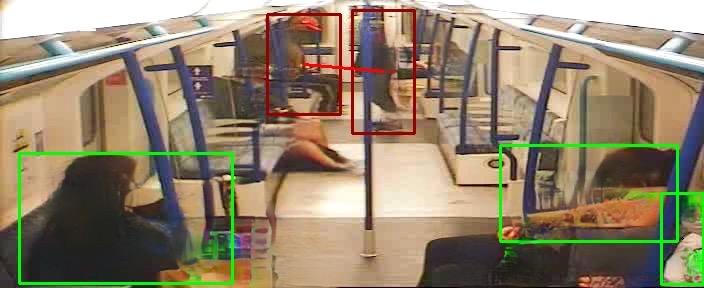}%
  \label{Badsd00}}
  \hfill
  \subfloat[A passenger is stretching out the leg.]{\includegraphics[width=0.35\textwidth]{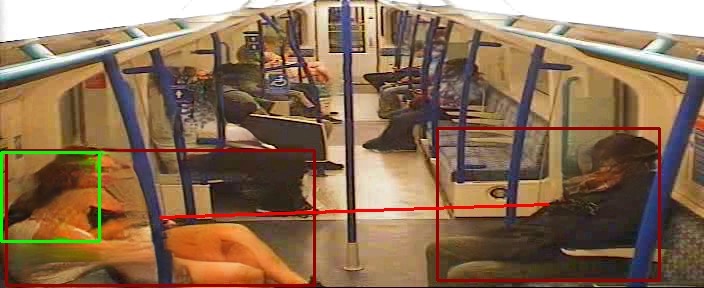}%
  \label{Badsd01}}
  \hfill
  \subfloat[A false positive of a seat next to a detected passenger.]{\includegraphics[width=0.35\textwidth]{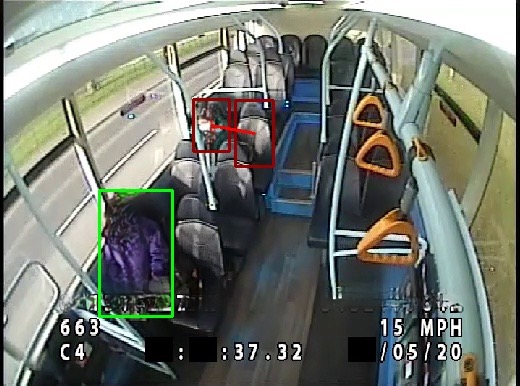}%
  \label{Badsd02}}
  \hfill
  \subfloat[The lower body of the passenger is not detected.]{\includegraphics[width=0.35\textwidth]{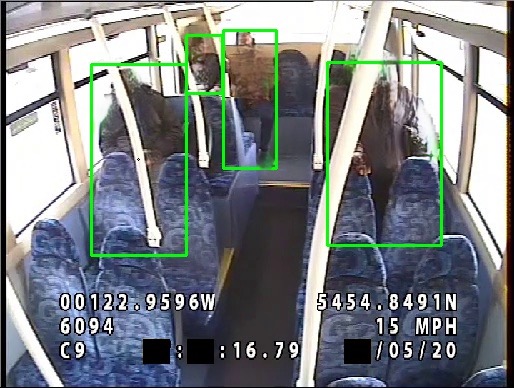}%
  \label{Badsd03}}

  \caption{Example of the incorrect social distancing measurement on the public transportation footages.}
  \vskip -10pt
  \label{fig:Badsd}
\end{figure}

\subsection{Social Distancing}
\label{sec:results_distance}

Using bounding boxes from the detected passengers, we can obtain the centroid of the boxes and measure the Euclidean distances of any adjacent passengers. We use this information and the safe distance threshold to measure the social distancing between passengers. We used Density-Based Spatial Clustering of Applications with Noise (DBSCAN) as density-based clustering where it clusters based on the distance between the nearest point \cite{ester1996density}. 

Figure \ref{fig:SD_train} shows the example of a social distancing measurement on public transportation. The green bounding box indicates safe social distancing, amber indicates not safe, and the red bounding box indicates dangerous social distancing. As the social distancing relies on passenger detection, the passenger that has not been detected did not contribute to the overall clustering algorithm, causing inaccurate measurement of the social distancing. It is also similar to any false positives that add inaccurate bounding boxes that will affect the accuracy of the social distancing measurement. 

It is also important to note that this social distancing measure is purely based on the passenger detection based on the footage. There is not enough information to conclude whether the passengers are perhaps from the same household.


The occlusion or the wrong pixel detection during passenger detection causes the wrong size of the bounding box, leading to inaccurate social distancing measurement as the clustering is based on the centre of the bounding box.

Some of the inaccurate social distancing measurement examples can be seen in Figure \ref{fig:Badsd} that shows a part of a passenger has been detected. However, the size of the bounding box is smaller than the actual ground truth, making the passenger's centroid a bit far from the next passenger. Contrary, as the passenger in Figure \ref{Badsd01} stretched out the legs, the bounding box became bigger and pushed the centroid closer to the other passenger. 

A false positive is detected next to a detected passenger in Figure \ref{Badsd02}, resulting in a false danger social distancing measurement. The seats on the vehicle can sometimes block some part of passengers resulting in a smaller bounding box than the actual ground truth. The two passengers sitting next to each other in Figure \ref{Bad03} are measured as at a safe distance because of the big difference in the bounding box size.
\section{Conclusion}

With the recent progress in machine learning for computer vision, there is much literature investigating the best way to tackle the issue of social distancing through the use of surveillance cameras in public spaces, primarily because of the COVID-19 pandemic. However, there is no related study on social distancing on public transportation up to date. CNNs have been shown capable in various object detection tasks within computer vision, especially when the image quality is good. In this work, we highlight the complexity of applying social distancing measurement using these methods on onboard low-resolution CCTV from public transportation. Using domain knowledge, we analyse the detection output to see if it is possible to improve the detection, especially in the headcount. Finally, we measure the social distancing between passengers using density-based clustering. 

In order to accurately monitor social distancing on public transportation, object detection should be able to detect every passenger on board by using the surveillance CCTV system on the vehicle. The physical location of the camera on public transportation, typically distant from the passenger seat location causes some passengers to appear as only a few pixels in the image.  Additionally, the nature of the public transportation with obstacles such as seats, handrails and other passengers may occlude clear line of sight to the passengers. In addition,  inconsistent image illumination and partial scenes captured from outside of the vehicles can be a potential issue degrading model predictive performance. 

The current onboard surveillance cameras on public transportation hinders satisfactory object detection and, consequently, affects the social distancing measurement. Other training approaches may be required to help models account for the challenges of these real-world images. Additionally, better quality hardware and improved physical placement of the camera should be put in place to improve automatic social distance measurement.



\section*{ACKNOWLEDGMENT}
We thank the UK Research and Innovation (UKRI), Department of Transport (DfT), Transport for London (TfL) and Go North East.

\bibliographystyle{IEEEtran}
\bibliography{ref}

\begin{thebibliography}{10}
\providecommand{\url}[1]{#1}
\csname url@samestyle\endcsname
\providecommand{\newblock}{\relax}
\providecommand{\bibinfo}[2]{#2}
\providecommand{\BIBentrySTDinterwordspacing}{\spaceskip=0pt\relax}
\providecommand{\BIBentryALTinterwordstretchfactor}{4}
\providecommand{\BIBentryALTinterwordspacing}{\spaceskip=\fontdimen2\font plus
\BIBentryALTinterwordstretchfactor\fontdimen3\font minus
  \fontdimen4\font\relax}
\providecommand{\BIBforeignlanguage}[2]{{%
\expandafter\ifx\csname l@#1\endcsname\relax
\typeout{** WARNING: IEEEtran.bst: No hyphenation pattern has been}%
\typeout{** loaded for the language `#1'. Using the pattern for}%
\typeout{** the default language instead.}%
\else
\language=\csname l@#1\endcsname
\fi
#2}}
\providecommand{\BIBdecl}{\relax}
\BIBdecl

\bibitem{saponara2021implementing}
S.~Saponara, A.~Elhanashi, and A.~Gagliardi, ``Implementing a real-time,
  ai-based, people detection and social distancing measuring system for
  covid-19,'' \emph{Journal of Real-Time Image Processing}, pp. 1--11, 2021.

\bibitem{khataee2021effects}
H.~Khataee, I.~Scheuring, A.~Czirok, and Z.~Neufeld, ``Effects of social
  distancing on the spreading of covid-19 inferred from mobile phone data,''
  \emph{Scientific Reports}, vol.~11, no.~1, pp. 1--9, 2021.

\bibitem{varghese2021multimodal}
E.~B. Varghese and S.~M. Thampi, ``A multimodal deep fusion graph framework to
  detect social distancing violations and fcgs in pandemic surveillance,''
  \emph{Engineering Applications of Artificial Intelligence}, vol. 103, p.
  104305, 2021.

\bibitem{kumari2021deep}
R.~Kumari and M.~U. Joshi, ``Deep learning and computer vision-based social
  distancing detection system,'' 2021.

\bibitem{magoo2021deep}
R.~Magoo, H.~Singh, N.~Jindal, N.~Hooda, and P.~S. Rana, ``Deep learning-based
  bird eye view social distancing monitoring using surveillance video for
  curbing the covid-19 spread,'' \emph{Neural Computing and Applications}, pp.
  1--8, 2021.

\bibitem{shalini2021social}
G.~Shalini, M.~K. Margret, M.~S. Niraimathi, and S.~Subashree, ``Social
  distancing analyzer using computer vision and deep learning,'' in
  \emph{Journal of Physics: Conference Series}, vol. 1916, no.~1.\hskip 1em
  plus 0.5em minus 0.4em\relax IOP Publishing, 2021, p. 012039.

\bibitem{punn2020monitoring}
N.~S. Punn, S.~K. Sonbhadra, S.~Agarwal, and G.~Rai, ``Monitoring covid-19
  social distancing with person detection and tracking via fine-tuned yolo v3
  and deepsort techniques,'' \emph{arXiv preprint arXiv:2005.01385}, 2020.

\bibitem{mercaldo2021proposal}
F.~Mercaldo, F.~Martinelli, and A.~Santone, ``A proposal to ensure social
  distancing with deep learning-based object detection,'' in \emph{2021
  International Joint Conference on Neural Networks (IJCNN)}.\hskip 1em plus
  0.5em minus 0.4em\relax IEEE, 2021, pp. 1--5.

\bibitem{hou2020social}
Y.~C. Hou, M.~Z. Baharuddin, S.~Yussof, and S.~Dzulkifly, ``Social distancing
  detection with deep learning model,'' in \emph{2020 8th International
  Conference on Information Technology and Multimedia (ICIMU)}.\hskip 1em plus
  0.5em minus 0.4em\relax IEEE, 2020, pp. 334--338.

\bibitem{DfTRail}
D.~of~Transport, ``Rail statistics,''
  \url{https://assets.publishing.service.gov.uk}, 2020.

\bibitem{morawska2020can}
L.~Morawska, J.~W. Tang, W.~Bahnfleth, P.~M. Bluyssen, A.~Boerstra,
  G.~Buonanno, J.~Cao, S.~Dancer, A.~Floto, F.~Franchimon \emph{et~al.}, ``How
  can airborne transmission of covid-19 indoors be minimised?''
  \emph{Environment international}, vol. 142, p. 105832, 2020.

\bibitem{zou2019object}
Z.~Zou, Z.~Shi, Y.~Guo, and J.~Ye, ``Object detection in 20 years: A survey,''
  \emph{arXiv preprint arXiv:1905.05055}, 2019.

\bibitem{liu2020deep}
L.~Liu, W.~Ouyang, X.~Wang, P.~Fieguth, J.~Chen, X.~Liu, and
  M.~Pietik{\"a}inen, ``Deep learning for generic object detection: A survey,''
  \emph{International journal of computer vision}, vol. 128, no.~2, pp.
  261--318, 2020.

\bibitem{lin2014microsoft}
T.-Y. Lin, M.~Maire, S.~Belongie, J.~Hays, P.~Perona, D.~Ramanan,
  P.~Doll{\'a}r, and C.~L. Zitnick, ``Microsoft coco: Common objects in
  context,'' in \emph{European conference on computer vision}.\hskip 1em plus
  0.5em minus 0.4em\relax Springer, 2014, pp. 740--755.

\bibitem{rahim2021monitoring}
A.~Rahim, A.~Maqbool, and T.~Rana, ``Monitoring social distancing under various
  low light conditions with deep learning and a single motionless time of
  flight camera,'' \emph{Plos one}, vol.~16, no.~2, p. e0247440, 2021.

\bibitem{yang2021vision}
D.~Yang, E.~Yurtsever, V.~Renganathan, K.~A. Redmill, and
  {\"U}.~{\"O}zg{\"u}ner, ``A vision-based social distancing and critical
  density detection system for covid-19,'' \emph{Sensors}, vol.~21, no.~13, p.
  4608, 2021.

\bibitem{zuo2021reference}
F.~Zuo, J.~Gao, A.~Kurkcu, H.~Yang, K.~Ozbay, and Q.~Ma, ``Reference-free
  video-to-real distance approximation-based urban social distancing analytics
  amid covid-19 pandemic,'' \emph{Journal of Transport \& Health}, vol.~21, p.
  101032, 2021.

\bibitem{pooranam2021safety}
N.~Pooranam, P.~P. Sushma, S.~Sruthi, and D.~K. Sri, ``A safety measuring tool
  to maintain social distancing on covid-19 using deep learning approach,'' in
  \emph{Journal of Physics: Conference Series}, vol. 1916, no.~1.\hskip 1em
  plus 0.5em minus 0.4em\relax IOP Publishing, 2021, p. 012122.

\bibitem{su2021novel}
J.~Su, X.~He, L.~Qing, T.~Niu, Y.~Cheng, and Y.~Peng, ``A novel social
  distancing analysis in urban public space: A new online spatio-temporal
  trajectory approach,'' \emph{Sustainable Cities and Society}, vol.~68, p.
  102765, 2021.

\bibitem{rezaei2020deepsocial}
M.~Rezaei and M.~Azarmi, ``Deepsocial: Social distancing monitoring and
  infection risk assessment in covid-19 pandemic,'' \emph{Applied Sciences},
  vol.~10, no.~21, p. 7514, 2020.

\bibitem{Vahab2019applications}
A.~Vahab, M.~S. Naik, P.~G. Raikar, and S.~Prasad, ``Applications of object
  detection system,'' \emph{International Research Journal of Engineering and
  Technology (IRJET)}, vol.~6, no.~4, pp. 4186--4192, 2019.

\bibitem{ren2015faster}
S.~Ren, K.~He, R.~Girshick, and J.~Sun, ``Faster r-cnn: Towards real-time
  object detection with region proposal networks,'' \emph{Advances in neural
  information processing systems}, vol.~28, pp. 91--99, 2015.

\bibitem{girshick2014rich}
R.~Girshick, J.~Donahue, T.~Darrell, and J.~Malik, ``Rich feature hierarchies
  for accurate object detection and semantic segmentation,'' in
  \emph{Proceedings of the IEEE conference on computer vision and pattern
  recognition}, 2014, pp. 580--587.

\bibitem{he2017mask}
K.~He, G.~Gkioxari, P.~Doll{\'a}r, and R.~Girshick, ``Mask r-cnn,'' in
  \emph{Proceedings of the IEEE international conference on computer vision},
  2017, pp. 2961--2969.

\bibitem{redmon2016you}
J.~Redmon, S.~Divvala, R.~Girshick, and A.~Farhadi, ``You only look once:
  Unified, real-time object detection,'' in \emph{Proceedings of the IEEE
  conference on computer vision and pattern recognition}, 2016, pp. 779--788.

\bibitem{redmon2018yolov3}
J.~Redmon and A.~Farhadi, ``Yolov3: An incremental improvement,'' \emph{arXiv
  preprint arXiv:1804.02767}, 2018.

\bibitem{wu2019detectron2}
Y.~Wu, A.~Kirillov, F.~Massa, W.-Y. Lo, and R.~Girshick, ``Detectron2,''
  \url{https://github.com/facebookresearch/detectron2}, 2019.

\bibitem{ester1996density}
M.~Ester, H.-P. Kriegel, J.~Sander, X.~Xu \emph{et~al.}, ``A density-based
  algorithm for discovering clusters in large spatial databases with noise.''
  in \emph{kdd}, vol.~96, no.~34, 1996, pp. 226--231.

\end{thebibliography}

\end{document}